\documentclass[acmlarge, screen]{acmart}
\usepackage{fancyhdr}
\fancypagestyle{plain}{%
  \fancyhf{} % Clear header/footer
  \fancyfoot[C]{\thepage} % Keep page numbers in footer
}
\usepackage{mathtools}
\usepackage{algorithm}
\usepackage{amsmath}
\usepackage{mathrsfs}
\usepackage{graphicx}
\usepackage{subcaption}
\usepackage{bbm}
\usepackage{natbib}
\usepackage{multirow}
\usepackage{algpseudocode}
\usepackage{wrapfig}
\usepackage{color}
\usepackage{svg}
\usepackage{xcolor}
\usepackage{tikz}
\usetikzlibrary{decorations.pathreplacing}

\usepackage[normalem]{ulem}
\definecolor{yg}{RGB}{127,0,255}
\definecolor{ORANGE}{RGB}{255, 165, 0}
\definecolor{RED}{RGB}{255, 0, 0}

\makeatletter
\renewcommand\paragraph{\@startsection{paragraph}{4}{\z@}%
                                    {3.25ex \@plus1ex \@minus.2ex}%
                                    {-1em}%
                                    {\normalfont\bfseries}}
\makeatother

\AtBeginDocument{%
  }

\setcopyright{acmlicensed}
\copyrightyear{2024}
\acmYear{2024}
\acmDOI{XXXXXXX.XXXXXXX}

\acmConference[Conference acronym 'XX]{Make sure to enter the correct
  conference title from your rights confirmation emai}
\acmISBN{978-1-4503-XXXX-X/18/06}

\theoremstyle{plain}

\theoremstyle{definition}

\theoremstyle{remark}

\newcommand{\ind}{\mathbbm{1}}
\newcommand{\var}{\mathrm{Var}}
\newcommand{\indep}{\rotatebox[origin=c]{90}{$\models$}}

\usepackage{xspace}

\definecolor{myblue}{rgb}{.8, .8, 1}
\definecolor{mathblue}{rgb}{0.2472, 0.24, 0.6} % mathematica's Color[1, 1--3]
\definecolor{mathred}{rgb}{0.6, 0.24, 0.442893}
\definecolor{mathyellow}{rgb}{0.6, 0.547014, 0.24}

\newcommand{\calC}{{\mathcal{C}}}
\newcommand{\calD}{{\mathcal{D}}}
\newcommand{\calE}{{\mathcal{E}}}

\newcommand{\calN}{{\mathcal{N}}}

\newcommand{\calS}{{\mathcal{S}}}
\newcommand{\calT}{{\mathcal{T}}}

\newcommand{\calX}{{\mathcal{X}}}
\newcommand{\calY}{{\mathcal{Y}}}

  \def\EE{\mathbb{E}}

  \def\PP{\mathbb{P}}
  
  \def\RR{\mathbb{R}}

\newcommand{\bfsym}[1]{\ensuremath{\boldsymbol{#1}}}

 \def\bOmega {\bfsym {\Omega}}

\def\vareps{\varepsilon}
\def\hat{\widehat}
\def\tilde{\widetilde}

\def\cal{\mathrm{cal}}

\DeclareTextFontCommand{\texttt}{\ttfamily\upshape}

\begin{document}

\title{Conformal Prediction: A Data Perspective}

\author{Xiaofan Zhou}
\affiliation{%
  \institution{University of Illinois Chicago}
  \city{Chicago}
  \state{IL}
  \country{USA}
}
\authornote{Both authors contributed equally to this research.}

\author{Baiting Chen}
\authornotemark[1]
\email{brantchen@g.ucla.edu}
\affiliation{%
  \institution{University of California-Los Angeles}
  \city{Los Angeles}
  \state{CA}
  \country{USA}
}

\author{Yu Gui}
\affiliation{%
  \institution{University of Chicago}
  \city{Chicago}
  \state{IL}
  \country{USA}}
\email{yugui@uchicago.edu}

\author{Lu Cheng}
\affiliation{%
  \institution{University of Illinois Chicago}
  \city{Chicago}
  \state{IL}
  \country{USA}
}
\email{lucheng@uic.edu}
% \affiliation{%

% \affiliation{%

% \affiliation{%

% \affiliation{%

% \affiliation{%


% motivation from data
\begin{abstract}
Conformal prediction (CP), a distribution-free uncertainty quantification (UQ) framework, reliably provides valid predictive inference for black-box models. CP constructs prediction sets or intervals that contain the true output with a specified probability. However, modern data science’s diverse modalities, along with increasing data and model complexity, challenge traditional CP methods. These developments have spurred novel approaches to address evolving scenarios. This survey reviews the foundational concepts of CP and recent advancements from a data-centric perspective, including applications to structured, unstructured, and dynamic data. We also discuss the challenges and opportunities CP faces in large-scale data and models.

\end{abstract}




\maketitle
% to do list
% Introduction
% conclusion
% notation table
%% method table
% writing
%% more figures
%% simplify 
% figure 1
\section{Introduction}

The recent rapid development of well-designed and powerful machine learning (ML) models has significantly transformed our lives. However, the success of these models is often evaluated based on the accuracy of their predictions, which, while important, is not sufficient in many real-world scenarios. In high-stakes applications, it is equally critical to assess the \textit{uncertainty} of model outputs. Uncertainty quantification (UQ) has long been a central problem in fields like statistics and ML. Several well-established methods, such as Bayesian inference and resampling techniques, have been widely adopted to address UQ. However, Bayesian posterior intervals are only valid if the distributional assumptions of the model are correctly specified, which may not always be the case in practical applications. Similarly, when constructing prediction sets for unobserved outputs, resampling approaches based on the residuals of observed data can result in overly narrow intervals since the estimate of variance is too small \citep{bates2024cross}, compromising the validity of the prediction set. This raises a fundamental question:
\begin{center}
\vspace{-0.5mm}
\textit{Is it possible to provide a valid confidence or prediction set,\\
regardless of the model in use and with minimal assumptions about the data distribution?}
\vspace{-0.5mm}
\end{center}
% More concretely, with a model $f(\cdot)$ taking the input $X$, we aim to construct a prediction set/interval $C(\cdot)$ that contains the underlying output $Y$ with a high probability, i.e. 
% \[
% \PP(Y \in C(X)) \geq 1 - \alpha,
% \]
% where $\PP$ is the probability with respect to the distribution of $(X,Y)$.
CP \cite{vovk2005algorithmic} is model-agnostic and can be applied to any ML model, generating prediction sets/intervals based on non-conformity scores derived from the model's outputs. CP offers several desirable features: (1) \textit{Finite-sample validity}: CP generates calibrated prediction sets/intervals that ensure high-probability coverage with finite-sample theoretical guarantees, without simplifying the models' complexity or relying on strong assumptions about the data generating process. More concretely, the validity of CP only requires the \emph{exchangeability} assumption of data. (2) \textit{Lightweight:} As a wrapper method, 
% split CP, a commonly adopted variant, 
CP does not require retraining the underlying model, making it computationally efficient and easy to apply to existing models. (3) \textit{Versatility:} CP is model-agnostic, meaning it can be used with any model, regardless of its architecture or complexity. (4) \textit{Informativeness:} When appropriately designed, CP's prediction set or interval can be sufficiently small to provide meaningful information about the original outputs.
\subsection{CP from a Data Perspective}
The increasing prevalence of big data, particularly with richer and more complex modalities, presents unique challenges for CP beyond traditional statistical settings. For instance: (1) \textbf{Generative models in natural language and image processing}: In these contexts, the target output can be ultra-high-dimensional  or even infinite. For example, the output of a language model can be a permutation of any tokens in the pre-training corpus with arbitrary length
, unlike the univariate real-valued responses in classical regression or classification tasks. For complex tasks like open-domain question answering, the output space is often unbounded \cite{su2024api,fischefficient} (i.e., all possible text sequences). Adapting CP to handle these multi-dimensional and even infinite-dimensional outputs is essential to improve its applicability and predictive accuracy. 
% {\color{red}(2) \textbf{Economics} is a common domain where machine learning and AI methods are widely applied, including in price prediction and fraud detection. However, the failure to achieve accurate predictions can result in significant costs, presenting substantial challenges for CP.} {\color{red}This example is weird, does Economics include data that traditional CP cannot address?} 
(2) \textbf{Time series and streaming data}: These types of data are becoming increasingly important in fields such as economics, meteorology, geosciences, urban planning, and healthcare. However, time-dependent data poses significant challenges for CP, particularly because the assumption of data exchangeability is often violated in the presence of distribution shifts or temporal dependencies. (3) \textbf{Structured data} (e.g., groups, trees, graphs): Structured data is typically quantitative and organized in predefined formats like rows and columns in relational databases while unstructured data (like text) is often qualitative and lacking a fixed schema \cite{zhang2020combining,tang2014feature}. In ML applications involving structured data, canonical CP methods may fail to capture the intrinsic dependencies when the feature structure is ignored. Traditional inferential methods often rely on distributional assumptions specific to structured data. Thus, developing model-agnostic UQ tools that leverage implicit (or weighted) exchangeability within the data structure is crucial.

\begin{figure}%[!ht]
  \centering
  \includegraphics[width=\linewidth]{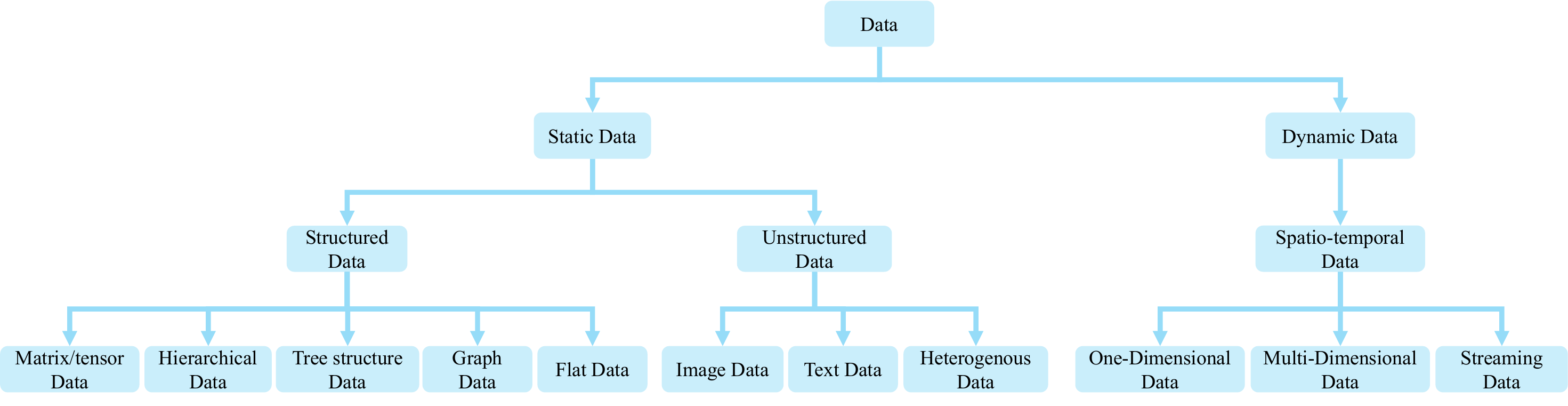}
  \caption{The proposed taxonomy for CP from a data perspective.}
  \label{fig:graph}
  \vspace{-9mm}
\end{figure}

These challenges motivate this survey on CP \emph{from a data perspective}, which introduces CP methods tailored to different data categories, as outlined in Figure \ref{fig:graph}. We categorize data into two main types based on the collection process: static data and dynamic data. Static data is further divided into structured data---including matrix/tensor data, hierarchical data, tree-structured data, and graph data---and unstructured data including image, text, and multi-modal data. Dynamic data is focused on spatio-temporal data, which is further subdivided into streaming, one-dimensional, and multi-dimensional categories. Note that dynamic data can be also structured, e.g., time-series data. This hierarchical organization provides a clear and systematic framework for understanding various data types, allowing for the appropriate application of CP techniques.

\subsection{Comparison with Existing Surveys}
There are several existing surveys that either cover the theoretical foundations of CP with a review of more advanced algorithm developments (e.g. \cite{angelopoulos2021gentle}) or review CP methods for a specific data type (e.g., \cite{sun2022conformal}). Particularly, \cite{angelopoulos2021gentle}, \cite{fontana2023conformal} and \cite{angelopoulos2024theoretical} introduce theoretical foundations of CP including split CP, full CP, conformalized quantile regression, as well as recent advanced applications of CP. The survey paper \cite{sun2022conformal} focuses on the application of CP in the specific field with spatio-temporal data and \cite{campos2024conformal} centers around the field of natural language processing (NLP). \cite{lindemann2024formal} focuses CP's application in autonomous systems, dealing with dynamic data by having trajectory calibration data. However, in light of the increasingly richer data types and modalities, a comprehensive and structured overview of CP from a data perspective is urgently needed to comprehensively understand the unique challenges of CP adapted to various data types. In this survey, we propose a novel taxonomy of CP from a data perspective (Figure \ref{fig:graph}) and review CP approaches and applications for structured data, unstructured data, and spatio-tempoal data. We will start with a background introduction to basic CP and then provide a thorough discussion of CP across different tasks and data types, examining how the nature of the data influences the effectiveness of CP methods. Lastly, we summarize several open challenges and promising future directions that hope to spark future research in CP.

% \subsection{Notations}
% Here we show some notations used in our paper.

\section{Background of CP} 
% \lu{simplify this section?}
In this section, we first introduce basic CP approaches and important variants and then summarize common metrics in evaluating UQ methods. The major notations used throughout the survey are summarized in Table \ref{table:your_label} in the supplementary. 
% \yg{Readers who are familiar with the CP literature can jump to Section~\ref{sec:conv_data}.}

\subsection{Full CP}
\begin{figure}[tb]
    \centering
\includegraphics[width=\linewidth]{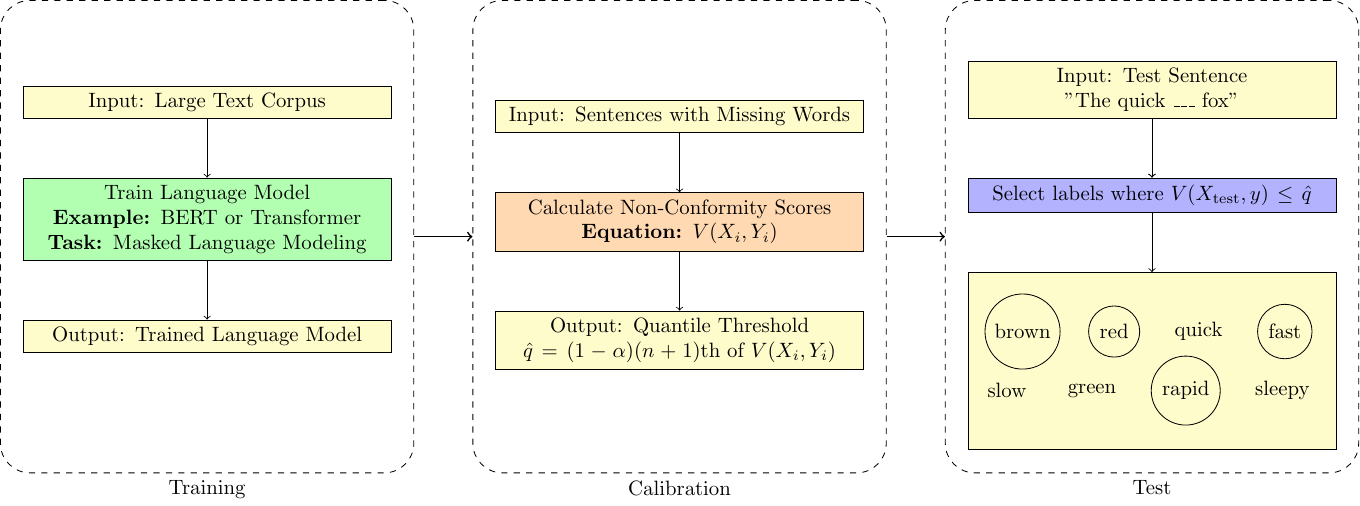}
    \caption{Example of Split CP for Text Infilling. In this task, the model predicts missing words in a sentence, and Split CP is used to create prediction sets that quantify the uncertainty of these predictions. \(V(X_i,Y_i)=1-P(Y_i|X_i).\) where \(P\) represents the predictive probability output by the trained base model.}
    \label{fig:cp_framework}
\end{figure}

% \lu{Since this is the first part with technical details, I would highly recommend to use examples (maybe fig 2) to illustrate the notations. For example, the score function $V$ is very new to outsiders, it would be good to explain using examples. Maybe replace the current fig 2 with examples that can explain all CP methods in Sec 2?}
Let $X$ and $Y$ denote features and label, respectively. Taking the regression problem as an example, consider observed data $(X_1,Y_1), \dots, (X_n,Y_n)$ and the test data $(X_{n+1}, Y_{n+1})$ with $Y_{n+1}$ unobserved. Our goal is to construct a prediction set for $Y_{n+1}$. Suppose that $(X_i,Y_i)$, $i=1,\dots,n+1$, are exchangeable (a relaxed assumption of i.i.d.\footnote{ Note that exchangeability does not imply independence. For example, consider variables $X_i + \vareps$, $i \in [n]$, where $X_1,\dots,X_n$ are i.i.d. and $\vareps$ is a shared additive term with $\var(\vareps) > 0$. Then, $X_i+\vareps$'s are exchangeable but the dependence is not zero.}), i.e. the joint distribution remains unchanged under any permutation, we have the property such that for any score function $V: \calX \times \calY \rightarrow \RR$, 
\begin{equation}
\left\{V(X_{n+1},Y_{n+1}) \mid \{V(X_i,Y_i)\}_{i=1}^{n+1} = \{v_i\}_{i=1}^{n+1}\right\} \sim {\rm Uniform}(\{v_i\}_{i=1}^{n+1}),
\end{equation}
which implies that conditioning on the realizations of the $n+1$ scores $\{v_i\}_{i=1}^{n+1}$, the test score $V(X_{n+1},Y_{n+1})$ can be viewed as a draw from the uniform distribution over all the realized scores.  The score function, also referred to as the \textit{non-conformity score}, quantifies how "atypical" a prediction is relative to the true value. Then, denoting the $(1-\alpha)$th-quantile of the uniform distribution over $\{V(X_i,Y_i))\}_{i=1}^{n+1}$ by $Q_{1-\alpha}$,
based on the property of the uniform distribution, we have
\begin{align}\label{eq:exch_valid}
\PP\left(V(X_{n+1},Y_{n+1}) \leq Q_{1-\alpha}\right) \geq 1 - \alpha.
\end{align}
However, as $Y_{n+1}$ is not observed, consider plugging in $Y_{n+1} = y$ in model fitting. Then, we denote $V^y(X,Y)$ as the score function trained with $Y_{n+1}=y$ and $Q^y_{1-\alpha}$ as the $(1-\alpha)$ th-quantile of the uniform distribution over $\{V^y(X_{n+1},y)\} \cup \{V^y(X_i,Y_i))\}_{i=1}^{n}$. Consider the following set
\begin{equation}
C_{1-\alpha}(X_{n+1}) = \{y \in \calY:\;V^y(X_{n+1},y) \leq Q^y_{1-\alpha}\},
\end{equation}
which, according to Eq. \eqref{eq:exch_valid}, satisfies that $\PP(Y_{n+1} \in C_{1-\alpha}(X_{n+1})) = \PP\left(V^{Y_{n+1}}(X_{n+1},Y_{n+1}) \leq Q_{1-\alpha}\right) \geq 1 - \alpha$. This indicates that with a probability no less than $1-\alpha$, the response $Y_{n+1}$ is included in the constructed set.
A common choice of the score function is $V^y(X,Y) = |Y - f^y(X)|$, where $f^y(X)$ is some fitted estimate of $\EE[Y \mid X]$ trained with $Y_{n+1}=y$. In Fig. \ref{fig:cp_framework}, this process is demonstrated for a text infilling task without calibration. Here, the score function measures the model’s uncertainty in predicting missing words on training data itself, and the resulting prediction set captures multiple plausible completions while maintaining the required coverage guarantee.

\subsection{Split CP and Variants} 
However, as we may have seen from above, full CP generally requires model re-fitting for all the values $y \in \calY$, which is infeasible in general scenarios with $|\calY| = \infty$. For computational efficiency, various variants of full CP are proposed, as detailed below. %We use the nonconformity score $V(X,Y) = |Y - \hat \mu(X)|$ in this section.

\textbf{Split CP} One of the most common approaches in CP is \emph{split CP} \cite{papadopoulos2002inductive,lei2018distribution}, which begins by pre-fitting a model $f$ and subsequently assessing its efficacy using a separate calibration set of labeled examples \(\{(X_i, Y_i)\}_{i=1}^n\), with which we obtain nonconformity scores $\{V_i = V(X_i, Y_i)\}_{i=1}^n$. 
% e.g. $V(x,y) = |y - \hat \mu(x)|$. 

Central to this method is the foundational assumption of exchangeability and the independence between the training and calibration sets. 
Based on the scores, we can construct the prediction set 
\begin{equation}
\label{eq:split_CP_set}
C(X_{n+1}) = \left\{y \in \calY:\;V(X_{n+1}, y) \leq \hat Q_{1-\alpha}\right\},
\end{equation}
where the quantile $\hat Q_{1-\alpha}$ can be directly obtained from the nonconformity scores:
\begin{equation}
\hat{Q}_{1-\alpha} = \text{the } \lceil(1-\alpha)(n + 1)\rceil\text{th smallest } V_i.
\end{equation}
The coverage guarantee of split CP ensures that the true value of a new observation falls within the predicted interval with high probability, i.e. $\PP(Y_{n+1} \in C(X_{n+1})) \geq 1 - \alpha$. To illustrate this process, Figure \ref{fig:cp_framework} provides a concrete example of Split CP applied to a text infilling task. The figure highlights how the calibration process enables a trained model to compute nonconformity scores and subsequently generate prediction sets with coverage guarantees.

Split CP involves just a single model fitting step, which, while computationally efficient, can lead to lower accuracy.
% \lu{what does statistical efficiency mean?} \z { According to $1-\alpha \leq \text{coverage} \leq 1-\alpha+\frac{1}{n+1}$, the higher upper bound highlights the extra conservatism introduced due to the limited calibration data and may produce wider prediction intervals}
Due to the reduction of sample size. In contrast, full CP achieves high statistical efficiency by requiring an extensive number of model fittings \cite{lei2018distribution,angelopoulos2021gentle}. However, these two methods represent extremes on a spectrum of possibilities. There are intermediate techniques that balance statistical and computational efficiency in different ways. %For example, cross-CP and CV+/Jackknife+ both require only a limited number of model fittings while still utilizing all available data for both model training and calibration.

\textbf{CP with cross-validation} 
As an extension of split CP, \cite{vovk2015cross,vovk2018cross} proposed cross-CP, which involves splitting the training dataset into $K$ disjoint subsets, $S_1, \ldots, S_K$, each of equal size $m = n/K$, where $n$ is divisible by $K$. For each subset $S_k$, a regression model $f_{-S_k}$ is trained on the complement of $S_k$ in the dataset. These models are then used to predict the outcomes of the samples within their respective subsets, resulting in a set of cross-validation residuals, e.g.
\begin{align}\label{eq:cvCP}
V_{i}^{\text{CV}} = |Y_i - f_{-S_{k(i)}}(X_i)|,
\end{align}
where $k(i)$ identifies the subset containing instance $i$. Consider the text infilling task, the text corpus is split into 
\(k\) disjoint subsets of sentences with missing words. For each subset, a language model is trained on the other 
\(k-1\) subsets and then used to predict the missing words in the held-out subset. However, the finite-sample validity of the cross-conformal predictors \cite{vovk2015cross,vovk2018cross} at the level of $1-\alpha$ may not hold. To address this issue, \cite{barber2021predictive} proposed CV+ 
 prediction intervals unsymmetrical around the prediction for $X_{n+1}$ but have the finite-sample coverage at least $1-2\alpha$.
% for K-fold cross-validation by defining the CV+ prediction interval as:
% \begin{align}\label{eq:C-cv}
% \hat{C}^{\text{CV+}}_{n,K,\alpha}(X_{n+1}) = \left[ \hat{q}_{n,\alpha}^- \left\{ \hat{\mu}_{-S_{k(i)}}(X_{n+1}) - V^{\text{CV}}_i \right\}, \hat{q}_{n,\alpha}^+ \left\{ \hat{\mu}_{-S_{k(i)}}(X_{n+1}) + V^{\text{CV}}_i \right\} \right].
% \end{align}
% which is shown to have the coverage guarantee
% \[
% P\left( Y_{n+1} \in \hat{C}^{\text{CV+}}_{n,K,\alpha}(X_{n+1}) \right) \geq 1 - 2\alpha - \min \left\{ \frac{2(1 - 1/k}{n/k + 1}, \frac{1 - k/n}{K + 1} \right\} \geq 1 - 2\alpha - \sqrt{2/n}.
% \]

% For a new observation $X_{n+1}$, the cross-CP method calculates prediction intervals by utilizing these residuals:
% \[
% \hat{C}_{n,K,\alpha}^{\text{cross-conf}}(X_{n+1}) = \left\{ y \in \mathbb{R} : \frac{\tau + \sum_{i=1}^n  \mathbf{1}\left[|y - \hat{\mu}_{-S_k(i)}(X_{n+1})| \leq R_i^{\text{CV}}\right] + U_i }{n+1} > \alpha \right\},
% \]
% where $U_i$ is a randomized term to break the tie.
% Here \(\tau \sim \text{Unif}[0,1]\) introduces randomization into the method.

\textbf{CP with jackknife} Jackknife-based CP can be viewed as a specific example of cross-validated CP with $K = n$. \citet{barber2021predictive} propose CP with jackknife+, which relieves the overfitting of canonical jackknife intervals and exhibits a finite-sample guarantee. Concretely, with leave-one-out (LOO) predictors $f_{-i}$, $i \in [n]$, one can calculate the residuals $V_i^{\text{LOO}} = |Y_i - f_{-i}(X_i)|$ and construct the prediction interval similar with Eq. \eqref{eq:cvCP} by defining $k(i) = i$. In the text infilling task, Jackknife+ trains a language model by leaving out one sentence at a time, computes residuals from the model’s predictions, and then uses these residuals to generate prediction sets for new inputs.
This modification is crucial in scenarios where predictions are sensitive to slight data variations, providing intervals that more accurately reflect the model's uncertainty. The Jackknife+ prediction interval satisfies the following coverage guarantee:
\begin{equation}
\mathbb{P} \{ Y_{n+1} \in \hat{C}_{n,1-\alpha}^{\text{jackknife+}}(X_{n+1}) \} \geq 1 - 2\alpha.
\end{equation}
Moreover, jackknife-minimax builds on the foundation of Jackknife+ by adopting a more conservative approach, aiming to remove the factor of two from the theoretical error bound to achieve tighter coverage control as is shown in \cite{barber2021predictive}. Tigher coverage guarantees around $1-\alpha$ are also presented in \cite{barber2021predictive} when algorithmic stability is assumed. Alternatives to relieve the computational costs of full CP are also explored in the literature. For example, \cite{martinez2023approximating} proposes approximate full CP and utilizes influence functions to construct easier-to-compute nonconformity scores.

% Jackknife-minimax builds on the foundation of Jackknife+ by adopting a more conservative approach, aiming to remove the factor of two from the theoretical error bound to achieve tighter coverage control \cite{barber2021predictive}:
% \[
% \hat{C}_{n,\alpha}^{\text{jack-mm}}(X_{n+1}) = \left[ \min_{i=1,\ldots,n} \hat{\mu}_{-i}(X_{n+1}) - \hat{q}_{n,\alpha}^{+} \{R_i^{\text{LOO}}\}, \max_{i=1,\ldots,n} \hat{\mu}_{-i}(X_{n+1}) + \hat{q}_{n,\alpha}^{+} \{R_i^{\text{LOO}}\} \right].
% \]
% This method ensures a coverage probability close to \(1-\alpha\), providing a more conservative interval that is crucial for applications requiring stringent reliability.
% The Jackknife-minimax prediction interval meets the following coverage requirement:
% \[
% \mathbb{P}\left(Y_{n+1} \in \hat{C}_{n,\alpha}^{\text{jack-mm}}(X_{n+1})\right) \geq 1 - \alpha.
% \]

\subsection{Weighted CP}
The exchangeability assumption is crucial in ensuring the finite-sample guarantee of CP.
However, in many practical scenarios with population heterogeneity or distribution shift, this assumption can be violated. \citet{tibshirani2019conformal} extended the principle of exchangeability to a weighted setting, providing a theoretical foundation for weighted CP when the data are weighted exchangeable. More concretely, consider the setting with covariate shift:
\begin{align}
(X_i, Y_i) \overset{\text{i.i.d.}}{\sim} P = P_X \times P_{Y|X}, \quad i \in [n], \qquad (X_{n+1}, Y_{n+1}) \sim \tilde{P} = \tilde{P}_X \times P_{Y|X}.
\end{align}
The data are weighted exchangeable in the sense that, with the weight function \(w(x) = \frac{d\tilde{P}_X}{dP_X}(x)\), 
\begin{equation}
\bigg(V(X_{n+1},Y_{n+1}) \mid \{V(X_i,Y_i)\}_{i \in [n+1]}\bigg) \sim \sum_{i = 1}^{n+1} \frac{w(X_i)}{\sum_{j=1}^{n+1} w(X_j)} \delta_{V(X_i,Y_i)},
\end{equation}
where $\delta$ is the point mass function. This means that given the observed values \(\{V(X_i, Y_i)\}_{i \in [n+1]}\), the new data point \((X_{n+1}, Y_{n+1})\) is considered to be drawn from a weighted combination of the empirical distribution of the observed data. 
% Then, taking the split CP framework as an example, one can construct a valid prediction interval $C(X_{n+1}) = [\hat \mu(X_{n+1}) - \hat{q}^w_{1-\alpha}, \hat \mu(X_{n+1}) + \hat{q}^w_{1-\alpha}]$ based on the weighted quantile
% \[
% \hat{q}^w_{1-\alpha} = {\rm Quantile}\bigg(1-\alpha; \sum_{i = 1}^{n} \frac{w(X_i)}{\sum_{j=1}^{n+1} w(X_j)} \delta_{V(X_i,Y_i)} + \frac{w(X_{n+1})}{\sum_{j=1}^{n+1} w(X_j)} \delta_{\infty}\bigg).
% \]
Weighted CP has been used across various fields such as experimental design \cite{fannjiang2022conformal}, survival analysis \cite{candes2023conformalized}, and causal inference \cite{lei2021conformal}. Subsequent studies have primarily focused on tackling the challenge of accurately estimating high-dimensional likelihood ratios to ensure effective coverage in complex data settings. For instance, \citet{barber2023conformal} extend it to general non-exchangeable settings, \citet{bhattacharyya2024group} enhance coverage guarantees in scenarios where observations belong to distinct groups, and \citet{ying2024informativeness,liu2024multi,lu2021distribution} bolster the informativeness of weighted CP when facing covariate shifts across multiple data sources.  To address the problem with distribution shift, there is another line of research focusing on calibrating the confidence level $\alpha$ to maintain validity for the worst-case distribution \citep{cauchois2024robust,ai2024not,gui2024distributionally}, which is known as distributionally robust methods.

\subsection{Conformal Risk Control}
% \lu{This part reads very different from the other subsections in Sec 2. Starts with motivation, and reduce the use of math notations. Use plain text to explain the key idea.}

Conformal risk control \cite{angelopoulosconformal} is a more general framework compared to traditional CP. While CP focuses on constructing prediction sets with guaranteed coverage probability, conformal risk control expands this idea to allow for more flexible risk management, considering different risk functions. Given a parametrized family of prediction sets $\{C_{\lambda}(x):\lambda \in \Lambda\}$, e.g. $C_{\lambda}(x) = \{y \in \calY: V(x,y) \leq \lambda\}$ with a given nonconformity score function $V$, the problem of constructing a valid prediction set boils down to the calibration of the parameter $\lambda$. The main idea is to choose $\lambda$ such that the average miscoverage in the calibration set is controlled, i.e. 
\begin{equation}
\hat \lambda = \inf \left\{\lambda \in \Lambda: \frac{1}{n_{\mathrm{cal}}+1}  \sum_{i=1}^{n_{\mathrm{cal}}} \ind\{Y_i \notin C_{\lambda}(X_i)\} + \frac{1}{n_{\mathrm{cal}}+1} \leq \alpha \right\}.
\end{equation}
Then, the prediction set $C_{\hat \lambda}(x)$ satisfies that $\PP(Y_{n+1} \in C_{\hat \lambda}(X_{n+1}) \geq 1-\alpha.$
Extensions to the generic loss function $\ell(x,y)$ are also presented in \cite{angelopoulosconformal}. 
Similar ideas are also explored in \cite{park2019pac,park2020pac,park2021pac}, in which PAC-type control of coverage is established.
% More applications of this procedure will be introduced in the following sections.

\subsection{Evaluation Metrics}
% \lu{Simplify it?}
In this section, we present a summary of common validity and accuracy guarantees for CP under a supervised setting. We denote $\hat C_{n,1-\alpha}$ as the prediction set based on the previous $n$ samples $\{(X_i,Y_i)\}_{i \leq n}$ at the level of $1-\alpha$ and consider the following metrics.

\subsubsection{Marginal coverage}
The most basic guarantee for a prediction interval is the finite-sample marginal coverage such that for the test sample $X_{n+1}$, one has
\begin{equation}
\PP(Y_{n+1} \in \hat C_{n,1-\alpha}) \geq 1 - \alpha,
\end{equation}
where the probability is with respect to the randomness of $\{(X_i,Y_i)\}_{i \leq n+1}$.
Furthermore, to obtain tight coverage, it usually requires that $\left| \PP(Y_{n+1} \in \hat C_{n,1-\alpha}) - (1 - \alpha) \right| = O(n^{-1})$.
CP is shown to meet both requirements with exchangeable data \cite{vovk2005algorithmic, shafer2008tutorial, lei2015conformal, alaa2024conformal}.

% Moreover, in \cite{vovk2002line}, under the exchangeability assumption, CP is also shown to satisfy the joint marginal validity in the online setting. More concretely, denote $\texttt{err}_i = \ind\{Y_{i+1} \notin \hat C_{i,1-\alpha}\}$. Then, for any $m \in \NN$ and any $\be \in \{0,1\}^m$,
% \[
% \PP\left\{(\texttt{err}_m, \cdots, \texttt{err}_1) = \be\right\} \leq \PP\left\{B^m(\alpha) = \be\right\},
% \]
% where $B^m(\alpha)$ is a vector of $m$ independent Bernoulli variables with mean $\alpha$.

\subsubsection{Adversarial and instantaneous coverage in online setting}
In the online setting with spatio-temporal data,  where data arrive sequentially and are not guaranteed to have the same distribution (see more details in Section~\ref{sec:spatio-temp}), the exchangeability assumption is violated, thus the worst-case or adversarial coverage is considered in the literature. 

\textbf{Adversarial marginal coverage.}
The worst-case coverage is the \textit{mean prediction interval coverage} averaged over all time points, which can be written as $T^{-1} \sum_{t=1}^{T} \ind\{Y_t \in \hat C_t(X_t)\}$,
where $T$ is the total number of time points and $\hat C_t(\cdot)$ is the prediction interval constructed at time $t$, and 0 otherwise \cite{xu2023sequential,zaffran2022adaptive,gibbs2021adaptive,yang2024bellman,stankeviciute2021conformal,barber2023conformal,jensen2022ensemble,xu2024conformal,xu2021conformal,sun2023copula,sousa2022general,lin2022conformalb, bhatnagar2023improved}.

\textbf{Instantaneous coverage.}
The worst-case marginal validity holds without distributional assumption for Adaptive conformal Inference (ACI)-type approaches, or alternatively, can be viewed as the property of a fixed sequence. 
To show the adaptivity of online CP approaches, \cite{angelopoulos2024online} also validates that the ACI-type has instantaneous validity under the i.i.d. assumption with sufficiently large $t$. With a fitted nonconformity score $s_t$, \cite{angelopoulos2024online} consider the asymptotic control of the coverage rate at time $t$: $\text{Coverage}_t(q) = \mathbb{P}(V_t(X, Y) \leq q \mid V_t).$

% where the probability is calculated with respect to a sample $(X, Y) \sim P$ drawn independently of $s_t$. 
% Here, $s_t$ is a score function at time $t$, $q$ is a quantile level, and $P$ represents the distribution of the sample. With $q_t$ updated at each step via $q_t = q_{t-1} + \eta_t (\text{err}_{t-1} - \alpha)$ with $\sum_{t \in \NN} \eta_t = \infty$ and $\sum_{t \in \NN} \eta_t^2 < \infty$, \cite{angelopoulos2024online} establishes the coverage guarantee such that $\text{Coverage}_t(q_t) \rightarrow 1-\alpha$ almost surely.

\subsubsection{Conditional coverage}\label{sec:cond-cov}
Conditional coverage is a stronger validity criterion. For example, if we are interested in a subpopulation determined by a function $g(x,y)$, it leads us to prediction intervals satisfying the conditional coverage guarantee such that for all $\tilde g$,
\begin{equation}
\PP\left(Y_{n+1} \in \hat C_{n,1-\alpha}(X_{n+1}) \mid g(X_{n+1},Y_{n+1}) = \tilde g\right) \geq 1-\alpha.
\end{equation}
In the following subsections, we consider different choices of the function $g$ in practice.

\textbf{Covariate-conditional coverage.}
A natural partition of the entire space is according to the realization of the test feature, i.e. $g(x,y) = x$, in which case we seek the covariate-conditional (or local) coverage.
% : for every $x \in \calX$,
% \[
% \PP\left(Y_{n+1} \in \hat C_{n,1-\alpha}(X_{n+1}) \mid X_{n+1} = x\right) \geq 1-\alpha.
% \]
Unfortunately, when the marginal distribution of $X$ has no point mass, this guarantee is shown to be impossible in \cite{vovk2005algorithmic,lei2014distribution}.
Various kinds of relaxation of the exact conditional coverage are explored in the literature (see \cite{foygel2021limits,guan2023localized,hore2023conformal,romano2019conformalized,lei2018distribution}). 

% \cite{foygel2021limits} considers a class of subsets $\calB = \{B: B \in \calX\}$ and relaxes the point-conditional validity to set-conditional validity in the sense that for all $B \in \calB$ with $\PP(X_{n+1} \in B) \geq \delta$,
% \[
% \PP\left(Y_{n+1} \in \hat C_{n,1-\alpha}(X_{n+1}) \mid X_{n+1} \in B\right) \geq 1-\alpha.
% \]
% It is shown in \cite{foygel2021limits} that when the VC dimension of $\calB$ is much smaller than $\delta n /\log^2(n)$, the conditional validity above can be achieved by a variant of split CP. However, when the VC dimension of $\calB$ is no less than $2n+2$, the relaxed conditional validity is still hopeless. \cite{guan2023localized} considers the localized modification of CP and \cite{hore2023conformal} further proposes a randomized variant to achieve the conditional guarantee. An asymptotic validity result is established in \cite{guan2022prediction} while \cite{hore2023conformal} provides a finite-sample guarantee with a coverage gap that explicitly measures the hardness of the set $B$ under the probability measure of $X$. Local performance of CP is also explored empirically in \cite{romano2019conformalized,lei2018distribution}.

\textbf{Class-conditional coverage.}
In the task of classification, it is of interest to provide valid prediction interval conditioning on each class, i.e. $g(x,y) = y$, in which we consider the label conditional guarantee.
% for each $k \in [K]$,
% \[
% \PP\left(Y_{n+1} \in \hat C_{n,1-\alpha}(X_{n+1}) \mid Y_{n+1} = k\right) \geq 1-\alpha.
% \]
In addition to the Mondrian CP in \cite{vovk2005algorithmic}, \cite{ding2024class} archives class-conditional validity in the setting where the number of classes is large such that effective samples in each class may not be sufficient. Empirical studies are conducted in \cite{lofstrom2015bias,shi2013applications,sun2017applying,hechtlinger2018cautious,guan2022prediction,sadinle2019least}.

\textbf{Training-conditional coverage.}
% Recall that the marginal coverage focuses on the performance averaged over both the training and test data. 
Since the coverage probably has high variability in the training data, it drives us to consider the coverage guarantee conditional on the training set $\calD_n$:
\begin{equation}
\PP\left(\alpha(\calD_n) \geq \alpha + \varepsilon_n\right) \leq \delta_n, \text{ where }
\alpha(\calD_n) = \PP\left(Y_{n+1} \notin \hat C_{n,1-\alpha}(X_{n+1}) \mid \calD_n\right),
\end{equation}
and $\varepsilon_n, \delta_n \rightarrow 0$ as $n \rightarrow +\infty$.
Specifically, the marginal coverage can be written as $\EE[\alpha(\calD_n)] \leq \alpha$.

% For split CP with $|\calD_n| = n_{\rm train} + n_{\rm calib}$, \cite{vovk2012conditional} proves training-conditional validity. CP with CV+ is similarly shown in \cite{bian2023training} to have training-conditional validity. However, \citet{bian2023training} demonstrate that, without additional assumptions, both full CP and CP with jackknife+ cannot guarantee training-conditional coverage. More recently, \citet{liang2023algorithmic} validate that algorithmic stability assumptions ensure training-conditional validity for CP.

\subsubsection{Prediction efficiency}
The efficiency of the prediction set/interval measures its size or length $\|\hat C_{n,1-\alpha}\|$. It is also common to compare with an oracle prediction set $C^*_{1-\alpha}$ under some specified models, see e.g.,  \cite{lei2018distribution}.
% . For example, in , it is shown that when the model is well-specified, one has $|\hat C_{n,1-\alpha}(X) \Delta C^*_{1-\alpha}(X)| = o_{\PP}(1)$, where $A \Delta B$ is the symmetric difference between sets $A$ and $B$.
 More criteria to measure prediction efficiency are discussed in \cite{vovk2016criteria}. For example, one can evaluate the second-highest p-values or the sum of all p-values excluding the largest one, as these metrics reflect the distribution of confidence across potential labels.
In the online setting, the mean prediction interval width, rolling width, and average median length of the prediction intervals are considered. 
To address the issue of infinite-length intervals, it is suggested to measure the \textit{fraction of infinite-length prediction intervals} \cite{yang2024bellman} and analyze the \textit{75\%, 90\%, and 95\% quantiles of the set sizes} \cite{angelopoulos2024conformal} to understand the distribution of interval lengths better.  

\subsubsection{Hybrid metric}
There are also several metrics that consider validity and efficiency together. \cite{lin2022conformala} introduces the \textit{inverse coverage efficiency}, which is the average prediction interval length divided by the marginal coverage rate (the smaller, the better). 
\cite{jensen2022ensemble} considers the \textit{coverage width-based criterion (CWC)} that penalizes under- and overcoverage in a similar way. Additional metrics that aim at balancing validity and accuracy are explored in \cite{sousa2022general,auer2024conformal}.

\section{Static Data---Structured data}
\label{sec:conv_data}

\subsection{Flat Data}
We start with the scenario where data are numeric and assumed to be independently and identically distributed (i.i.d or \emph{flat}), which is common in regression/classification problems.
% As the i.i.d. assumption is a weaker version of exchangeability required for the validity of CP, it affords practitioners flexibility in the choice of nonconformity scores. In addition, variants of the coverage guarantee (e.g. conditional coverage, group-wise coverage) are further explored.

\subsubsection{Regression}
Consider observed covariates $\{X_i\}_{i=1}^n \subseteq \RR^d$ and response $\{Y_i\}_{i=1}^n \subseteq \RR$, where we assume $\{(X_i,Y_i)\}_{i=1}^n$ are i.i.d. Then, according to the previous section, for any nonconformity score $V(x,y)$, the prediction set defined in Eq. \eqref{eq:split_CP_set} is valid. Particularly, for the regression problem, the score $V(x,y) = |y - \hat f(x)|$ is widely adopted \citep{vovk2005algorithmic,shafer2008tutorial,lei2018distribution} with a fitted function $f$, with which the prediction interval takes the form 
$C_{1-\alpha}(X_{n+1}) = f(X_{n+1}) \pm \hat Q_{1-\alpha}.$ Note that the prediction interval has an equalized width for different $X_{n+1}$. To incorporate the local uncertainty of covariates when the noise distributions are heteroscedastic, a locally weighted score is proposed \citep{papadopoulos2002inductive,lei2018distribution}:
\begin{equation}
V(x,y) = \frac{|y - \hat f (x)|}{\hat \sigma(x)},
\end{equation}
where the local uncertainty score $\hat \sigma(x) > 0$ is an estimated standard error of the residual. Variants of the local uncertainty score are proposed and applied to various scenarios \cite{papadopoulos2011reliable,ho2010martingale,cortes2018deep}.

However, as the distribution of $(Y \mid X)$ may not lie in the location-scale family such that $Y = f(X) + \sigma(X) \cdot e$, the prediction interval based on the locally weighted score can lose efficiency when dealing with more complicated distributions.
Another notable line of research focuses on nonconformity scores based on more general distributional quantities such as estimated quantiles and cumulative density functions (CDF), which are introduced as follows.

\textbf{Conformalized quantile regression(CQR).}
 Even when the distribution does not belong to the location-scale family, the quantile function can capture the intrinsic characteristics (e.g. the shape) of the distribution.
\cite{romano2019conformalized} proposes CQR where the prediction interval is defined as
\begin{equation}
\left[\hat Q_{\alpha/2}(X_{n+1}) - \eta_{1-\alpha}, \;\hat Q_{1-\alpha/2}(X_{n+1}) + \eta_{1-\alpha}\right],
\end{equation}
where $\hat Q$ is the estimated quantile based on data splitting and on the calibration set with sample size $n_{\mathrm{cal}}$, $\eta_{1-\alpha}$ is the $(1-\alpha)(1+1/n_{\mathrm{cal}})$th-quantile of 
\begin{equation}
V_i = \max\left\{\hat Q_{\alpha/2}(X_i) - Y_i, \;Y_i - \hat Q_{1-\alpha/2}(X_i)\right\}, \qquad 1 \leq i \leq n_{\mathrm{cal}}.
\end{equation}
When the estimated quantile is accurate, CQR also exhibits good local performance and to improve local coverage, further extensions are proposed in \cite{sesia2020comparison,alaa2023conformalized,sousa2022improved,rossellini2024integrating}.

\textbf{Distributional CP.}
In comparison, with the same motivation to capture the characteristics of the distribution beyond the location-scale family and achieve better conditional performance, distributional CP \citep{chernozhukov2021distributional} considers the scores based on an estimated CDF. Consider the data splitting case where $\hat F(x,y)$ is an estimated CDF using a hold-out set. \cite{chernozhukov2021distributional} defines the scores as
\begin{equation}
V_i = \psi\left(\hat F(X_i,Y_i)\right), \qquad 1 \leq i \leq n_{\mathrm{cal}},
\end{equation}
where $\psi$ is a user-specified scalar function to enhance accuracy and the default choice is $\psi(t) = |t - 1/2|$. When the CDF is estimated accurately, distributional CP is also shown to have an asymptotic conditional coverage guarantee.

\textbf{Discussion.} CP with flat data in the regression setting mainly focuses on designing nonconformity scores to achieve better local coverage. Further discussion is presented in Section~\ref{sec:cond-cov}.

\subsubsection{Classification}
\label{class}
% \lu{this subsection is too long. A figure or table is needed to summarize it.}
To apply CP to classification problems with $K$ classes such that $Y \in [K]$, a widely adopted nonconformity score is the estimated likelihood \citep{shafer2008tutorial,sadinle2019least}. Taking split CP as an example where we have $\hat p(x;k)$ as an estimate for $\PP(Y=k \mid X = x)$, then with the non-conformity score $V_i = \hat p(X_i;Y_i)$, the prediction set takes the form
\begin{equation}
C_{1-\alpha}(X_{n+1}) = \left\{k:\;\hat p(X_{n+1};k) \geq \hat Q_{1-\alpha}\right\}, 
\end{equation}
where $\hat Q_{1-\alpha}$ is the $\lfloor \alpha \cdot (1+n_{\mathrm{cal}}) \rfloor$th-smallest of $\{V_i\}_{i=1}^{n_{\mathrm{cal}}}$. 

A modified approach is proposed in \cite{romano2020classification} where the nonconformity score takes the form
\begin{equation}
\tilde V_i = \sum_{k \in \calS_i} \hat p(X_i;k), \qquad \calS_i = \{k:\hat p(X_i;k) \geq \hat p(X_i;Y_i)\},
\end{equation}
which is a score that tends to be small for the more ``likely'' labels for a certain $x$.
Let $\hat Q_{1-\alpha}$ be the $\lceil (1-\alpha)(1+n_{\mathrm{cal}}) \rceil$th-smallest of $\{\tilde V_i\}_{i=1}^{n_{\mathrm{cal}}}$ and $\{\pi(j;x)\}_{j=1}^{K}$ be permutation of $[K]$ such that
\begin{equation}
\hat p(x;\pi(1;x)) \geq \hat p(x;\pi(2;x)) \geq  \cdots \geq \hat p(x;\pi(K;x)).
\end{equation}
Then, the prediction set takes the form 
\begin{equation}
C_{1-\alpha}(X_{n+1}) = \{\pi(k;X_{n+1}): 1 \leq k \leq k(X_{n+1})\},
\end{equation}
where $k(x) = \min\{k: \sum_{j=1}^k \hat p(x;\pi(j;x)) \leq \tilde Q_{1-\alpha}\}$ and $\tilde Q_{1-\alpha}$ is the $\lceil (1-\alpha)(1+n_{\mathrm{cal}}) \rceil$th-smallest of $\{\tilde V_i\}_{i=1}^{n_{\mathrm{cal}}}$. In \cite{cauchois2021knowing}, a nonconformity score based on estimated quantiles, which is similar to CQR, is adopted for better local coverage performance. More recent work \citep{huang2023conformal} further improves the proposed nonconformity score via label ranking. \cite{einbinder2022training} calibrates the output probability of deep neural networks for more informative prediction sets.
% Applications of CP in classification are deferred to the following sections regarding different data types.

\textbf{Ordinal classification.} When the labels are ordinal, the prediction set is required to be contiguous. However, the prediction sets constructed above for general classification problems do not have this property unless the estimated conditional distribution $\hat P_{Y \mid X}$ is unimodal. Motivated by this property, \cite{dey2024conformal} proposed a procedure that constructs unimodal classifiers. In addition, \cite{chakraborty2024conformalized} adopts conformal p-values to determine the lower and upper bounds for the contiguous prediction set for the ordinal responses, and \cite{xu2023conformal} applies the conformal risk control framework to this setting. Conformalized ordinal classification has applications in disease severity rating \citep{lu2022improving,dey2024conformal,chakraborty2024conformalized}, age recognition \citep{dey2024conformal,xu2023conformal}, and historical image dating \citep{dey2024conformal}, and so on.

% \z{ The following content seems a bit confusing to me. It doesn't appear to be directly related to classification. For example, "Approximate Full CP" seems more like a variant of full CP, and "classification with data noise" seems more related to the section on "data with noise". I suggest reorganizing the content with more informative titles, such as specific tasks in classification: label imbalance classification, multi-label classification, etc.}

\textbf{Multi-label classification.}
% For multi-label classification, the target of this method is training a method $f: \calX \rightarrow \calY, \calX\in R, Y\in \{-1, 1\}^K$, where is $K$ is the number of all labels, $Y=1$ represents $X$ has this label while $Y=-1$ represents $X$ do not have this label. When applying CP, we wish to output a list $y\in {-1,1}^K$ which contains the ground truth $y$ with probability at least $1 - \alpha$.
% The adjusted calibration threshold $\hat{\tau}_k$ incorporates the inflation factor to correct for the noise:
% \begin{align}
%     \hat{\tau}_k = \inf \left\{ t \in [0,1] : \frac{1}{n_k} \sum_{i=1}^{n_k} I(s_i \leq t) \geq 1 - \alpha - \hat{\Delta}_k(t) + \delta(n_k, n^*) \right\}
% \end{align} 
% \(\delta(n_k, n^*)\) is a finite-sample correction term to ensure valid coverage even with a limited number of samples.
Maxime Cauchois et al \cite{cauchois2021knowing} apply CP on multi-class and multi-label image classification. One key challenge here is that typical CP methods—which give marginal validity (coverage) guarantees—provide uneven coverage, in that they address easy examples at the expense of essentially ignoring difficult examples. They build tree-structured classiﬁers that eﬃciently account for interactions between labels. To asymptotically get conditional coverage for both multiclass and multilabel prediction, they adopt the idea from quantile regression \cite{romano2019conformalized}.

\textbf{Discussion.} Conditional coverage is also of concern in classification problems, which motivates a series of works \citep{romano2020classification,cauchois2021knowing} in constructing scores similar to conditional quantiles. In addition, UQ for ordinal classification with additional rank constraints on the labels, which lies between regression and classification with unordered labels, is still an active field. Selective classification is also of interest in ML, where more cautious prediction sets with the ability to abstain are constructed \cite{hechtlinger2018cautious}

\subsubsection{Multivariate response}
\begin{figure}[tb]
    \centering
    \includegraphics[width=0.8\linewidth]{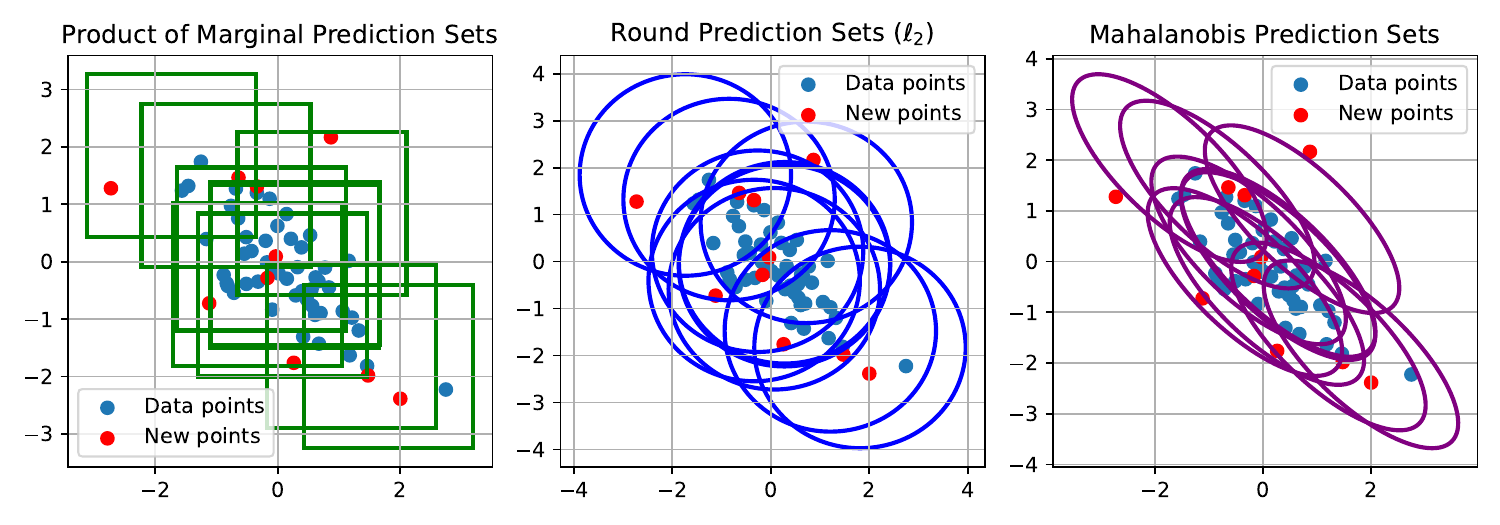}
    \caption{ Illustration of conformal prediction sets with bivariate response.}
    \label{fig:cp-biv}
\end{figure}

As an extension to CP with univariate response $Y$, the setting with a multivariate response is also of interest in the scenario where $Y \in \RR^k$ is a $k$-dimensional series for each data point. Although the validity of CP still holds with properly defined nonconformity scores, the specification of the scores is a crucial question to determine the shape of the prediction set for better accuracy. A naive approach is to construct the CP prediction interval $\hat C_j(X_{n+1})$ at the level of $1-\alpha_j$ for $Y_{n+1,j}$ for each $j \in [k]$ as is depicted in Figure~\ref{fig:cp-biv} (left panel). Then, by the union bound, one has $\PP\{Y_{n+1} \in \hat C_1(X_{n+1}) \times  \cdots \times \hat C_k(X_{n+1})\} \geq 1 - \sum_{j=1}^k \alpha_j$. Specifically, when coordinates in $Y$ are independent, we can further have $\PP\{Y_{n+1} \in \hat C_1(X_{n+1}) \times  \cdots \times \hat C_k(X_{n+1})\} \geq \prod_{j=1}^k (1-\alpha_j)$, which has much tighter overage.
To achieve a middle-ground between the independent and adversarially dependent settings, \cite{messoudi2021copula} and \cite{sun2023copula} further consider the copula transformation to achieve independent coordinates and apply an inverse map to obtain the predict set for the multivariate response. In comparison, \cite{johnstone2022exact} extends the score $|Y - \hat f(X)|$ to $\|Y - \hat f(X)\|$ with different choices of the norm $\|\cdot\|$ in $\RR^k$ and define the prediction set by $\{y \in \RR^k:\;\|y - \hat f(X_{n+1})\| \leq \hat q_{1-\alpha}\}$ with calibrated quantile $\hat q_{1-\alpha}$ for the scores. See Figure~\ref{fig:cp-biv} (middle and right panels) for prediction sets with two different choices of norms: $\|Y - \hat f(X)\|_2$ and $\|Y - \hat f(X)\|_{\bOmega} = (Y - \hat f(X))^\top \bOmega^{-1} (Y - \hat f(X))$, where $\bOmega$ is an estimate of the covariance matrix of residual vectors. Other extensions to the setting with multivariate responses are discussed in \cite{xu2024conformal,diquigiovanni2022conformal,feldman2023calibrated,dheur2024distribution}. 

\subsubsection{Functional data} 
CP with functional data can be viewed an a further extension to the previous section with multivariate responses since the response here is a function in an infinite-dimensional space.
Consider the setting with random functions $X_1(\cdot), \dots, X_n(\cdot) \in \Omega$ with domain $\calT$. The goal is to find a prediction set $\calC_{1-\alpha} \subseteq \Omega$ such that $\PP(X_{n+1} \in \calC_{1-\alpha}) \geq 1 - \alpha$. For brevity, we consider the split CP. To solve this problem, a naive trial is to find a predefined function $g: \Omega \rightarrow \RR$, e.g. $g(X) = - \sup_{t \in \calT}|X(t)|$, and then define $\calC_{1-\alpha} = \{x \in \Omega: g(x) \geq \lambda\}$. Here $\lambda$ is $(\lceil \alpha(n_{\mathrm{cal}}+1) \rceil - 1)$th-smallest among $\{g(X_i)\}_{i=1}^{n_{\mathrm{cal}}}$. However, this criterion is too ambitious and may lose efficiency. In \cite{lei2015conformal}, a modified criterion is considered:
$\PP(\Pi(X_{n+1}) \in \calC_{1-\alpha}) \geq 1-\alpha,$ where $\Pi$ maps functions in $\Omega$ to a finite dimensional space $\Omega_p \subset \Omega$. More concretely, \cite{lei2015conformal} considers the prediction bands $B(t) = [l(t), u(t)]$, $t\in \calT$, such that $\PP(l(t) \leq X_{n+1}(t) \leq u(t),\;\forall t\in \calT) \geq 1 - \alpha$. The proposed approach projects functions to vectors $\{\xi_i\}_{i \leq n} \subseteq \RR^p$ in the Euclidean space via a basis $\{\phi_j\}_{j=1}^p$.
Then, with nonconformity scores $ V(\xi_i)$, $i \in [n_{\mathrm{cal}}]$ and
\begin{equation}
\calE_n = \left\{\zeta \in \RR^p: V(\zeta) \geq  V_{(\lceil \alpha(n_{\mathrm{cal}}+1) \rceil - 1)}\right\},
\end{equation}
where $V_{(k)}$ is the $k$th order statistic of $\{V(\xi_i)\}_{i=1}^{n_{\mathrm{cal}}}$,
one can construct the prediction band by $B(t) = \left\{\sum_{j=1}^p \zeta_j \phi_j: \zeta \in \calE_n\right\}$.
In practice, the nonconformity score $V(\cdot)$ is chosen to be the Gaussian mixture density estimator with $K$ components.
\cite{diquigiovanni2021importance} constructs prediction bands for $X_{n+1}$ without projection by considering the nonconformity score
\begin{equation}
V(X_i) = \sup_{t \in \calT}\bigg|\frac{X_i(t) - \hat G(t)}{\hat s(t)}\bigg|, \quad i = 1, \dots, n_{\mathrm{cal}},
\end{equation}
where $\hat G(t)$ is the mean of $X_i(t)$ in the hold-out training set and $\hat s(t)$ is designed to penalize extreme values in \cite{diquigiovanni2021importance}. This procedure is further extended to the multivariate functional setting \cite{diquigiovanni2022conformal}.

\subsubsection{Data with noise, missingness, or censoring}
The ideal exchangeability assumption can be violated in many scenarios, among which data corruption is a common setting in statistics.
In this section, we summarize related works on CP with flat data but either the covariates or the response is not directly observed, which can be missing or perturbed.

\textbf{Covariate noise.}
The simplest setting with data corruption is the case with noisy covariates for the test data, which may happen when measurement errors are taken into account \citep{cochran1968errors} or perturbations are applied to enhance privacy \citep{dwork2006calibrating}.
Consider the case where we observe $\{(X_i,Y_i)\}_{i=1}^n$ and a new sample with noisy covariates $\tilde X_{n+1}$. \citet{gendler2021adversarially} adopt randomized smoothing to construct a nonconformity score $\tilde V(x,y)$ which satisfies that
\begin{equation}
V(\tilde X_{n+1},y) \leq V(X_{n+1},y) + M_{\delta}, \qquad \forall y \in \calY,\quad \text{almost~surely},
\end{equation}
where $M_{\delta}$ is a constant. Concretely, given any $V(x,y)$, the score $\tilde V(x,y)$ can be constructed by
\begin{equation}
\tilde V(x,y) = \Phi^{-1}\left(\EE_{\nu \sim \calN(0_d,I_d)}[V(x+\nu,y)]\right),
\end{equation}
where $\Phi$ is the CDF of $\calN(0_d,I_d)$. A more recent work \citep{ghosh2023probabilistically} only requires $V(\tilde X_{n+1},y) \leq V(X_{n+1},y) + M_{\delta}$ to hold with a high probability and proposes an adaptive approach to achieve improved trade-offs between performance on clean data and robustness to adversarial examples based on the principle of quantile-of-quantile design. \cite{zargarbashirobust} further considers the performance of CP under poisoning/evasion attack.

\textbf{Label noise.}
\cite{feldman2023conformal} focuses on the setting where we observe $\{(X_i,g(Y_i))\}_{i=1}^{n}$ with a corruption function $g(\cdot)$ and the goal is to construct a valid prediction interval for $Y_{n+1}$. For any $u \in (0, \alpha-1/(n+1))$, if the test score $V_{n+1} = V(X_{n+1},Y_{n+1})$ and the corrupted score $\tilde V_{n+1} = V(X_{n+1},\tilde Y_{n+1})$ satisfy
\begin{equation}
\PP(\tilde V_{n+1} \leq t) + u \leq  \PP(V_{n+1} \leq t),
\end{equation}
then with the protection interval $\tilde C_{1-\alpha}(\cdot)$ based on $(X_i,g(Y_i))$'s, it holds that
\begin{equation}
\PP(Y_{n+1} \in \tilde C_{1-\alpha}(X_{n+1})) \geq 1 - \alpha + \frac{1}{n+1} + u.
\end{equation}
Moreover, if the noisy distribution of $\tilde Y = g(Y)$ satisfies that $d_{\rm TV}(Y,\tilde Y) \leq \varepsilon$, one has
\begin{equation}
\PP(Y_{n+1} \in \tilde C_{1-\alpha}(X_{n+1})) \geq 1 - \alpha + \frac{1}{n+1} + \frac{n}{n+1}\varepsilon,
\end{equation}
which shows CP sets are robust when the strength of label noise is mild. In the study \cite{sesia2023adaptive}, they propose a novel CP method that adjusts for label contamination through a precise characterization of the effective coverage inflation or deflation caused by label contamination. Recently, a nonconformity score that interpolates noisy and clean data to achieve robustness is proposed \cite{penso2024conformal}. 

\textbf{Covariate missingness.}
In \cite{zaffran2023conformal}, the setting with missing coordinates in $X_i$'s is considered, which can be viewed as a specific case of noisy covariates. 
For example, the test data can take the form $X_{n+1} = (X_{n+1,1}, \cdot, \cdot, X_{n+1,4})$, where the second and third entries are unobserved.
Solutions are proposed in \cite{zaffran2023conformal} and the main idea is to artificially create missingness in the calibration dataset such that the second and third entries in calibration data are also missing. In the first approach, calibration data with additional missingness (e.g. $X_j = (X_{j,1},\cdot,\cdot,\cdot$) will be removed, in which case the exchangeability is recovered. To relieve the issue of sample size insufficiency, \cite{zaffran2023conformal} also suggests that we could impute $X_j$ to obtain $X_j = (X_{j,1},\cdot,\cdot,\hat X_{j,4})$ for CP.
Another line of work focusing on data (e.g. matrix or tensor) completion will be introduced in Section \ref{sec:other}.

\textbf{Causal inference.}
Causal inference under the potential outcome model \citep{rubin1974estimating} can be viewed as an example of the missing response. Consider covariates $X_i$'s and potential outcomes $(Y_i(0), Y_i(1))$ for each unit. A treatment $T_i \in \{0,1\}$ is assigned to unit $i$ and the strong ignorability assumption is commonly considered: $(Y_i(0),Y_i(1)) \indep T_i \mid X_i,$ i.e., no unobserved hidden confounders. The two major challenges of applying CP in causal inference are the covariate shift and inductive bias due to counterfactuals.
In \cite{lei2021conformal}, under the strong ignorability assumption, prediction intervals are constructed for random variables, such as $Y_{n+1}(1)$ and $Y_{n+1}(1) - Y_{n+1}(0)$, via weighted CP. To see this, for $Y_{n+1}(1)$, the density ratio between the population of interest $P_{Y(1),X \mid T=1}$ and $P_{Y(1) ,X}$ only depends on $X$, which lies in the covariate shift setting. Different choices of weight functions are presented in \cite{lei2021conformal} for various inferential targets. In comparison, \cite{kivaranovic2020conformal} adopts CP without reweighting to obtain prediction intervals that are valid conditioning on the treatment $T$. 
% \lu{Beside covariate shift, another challenge of applying CP in causal inferences} is the inductive bias, \lu{why?}. \citet{alaa2024conformal} propose conformal meta learner which combines CP with meta-learner \cite{kunzel2019metalearners}. Their method tackles the covariate shift because the distribution of covariates linked with pseudo-outcomes remains consistent between training and testing data. It also addresses inductive bias since the calibration step is independent of the model architecture, allowing for a flexible selection of inductive priors. This flexibility enables the reuse of existing meta-learners and architectures that have demonstrated accurate estimates of \lu{causal effects}.
\cite{jin2023sensitivity,chen2024conformal} propose robust CP that maintains valid coverage when the assumption is slightly violated. More concretely, for example, with the bounds constraint of the density ratios, the weight function can be replaced by a lower or an upper bound function in calculating the weighted quantile for weighted CP. In addition, \cite{yang2024doubly,qiu2023prediction,chen2024biased} further propose doubly robust results by incorporating the missingness/covariate shift in the framework of nonparametric/semi-parametric statistics and \citet{alaa2024conformal} propose conformal meta learner which combines CP with meta-learner \cite{kunzel2019metalearners} to estimate causal effects.

\textbf{Censored data.}
The setting with the censored response is commonly known as survival analysis where $X$ consists of covariates, $T$ is the survival time of interest and $C$ is the censoring time. In practice, only the censored time $\tilde T = \min\{C,T\}$ can be observed. With the goal of constructing a lower prediction bound (LPB) for $T$ such that $\PP(L_{1-\alpha}(X) \leq T) \geq 1-\alpha$, if we directly treat $\tilde T$ as the response, the LPB for $\tilde T$ is also valid for $T$. However, when $C < T$ with a high probability, the LPB for $\tilde T$ will be too conservative for $T$. The main idea in \cite{candes2023conformalized,gui2024conformalized} is to find a cutoff $q(X)$ for $C$ and focus on a subpopulation with $C \geq q(X)$, in which the censoring rate is lower. The difference lies in that \cite{candes2023conformalized} adopts a fixed cutoff $c_0$ while \cite{gui2024conformalized} considers an adaptive cutoff as a function of $X$. We note that, under the conditionally independent censoring assumption that $T \indep C \mid X$, the distribution shift from $(T,C,X)$ to $(T,C,X\mid C \geq q(X))$ is covariate shift. Thus, by weighted CP, the LPB for $\tilde T$ constrained on the subpopulation will gain more accuracy.

\textbf{Discussion.}
UQ for a wide range of statistical problems with missing/noisy observations can be addressed via (weighted) CP. However, its validity usually depends on the covariate shift condition, which has different variants in specific problems (e.g. the strong ignorability assumption in causal inference and the conditionally independent censoring assumption in survival analysis). Motivated by \cite{jin2023sensitivity,barber2023conformal}, there is ongoing interest in robustness evaluation or sensitive analysis of (weighted) CP when the covariate shift assumption is violated.

\subsection{Other Structured Data}
\label{sec:other}
The validity of CP holds with flat data. However, in practice, the i.i.d. assumption is too strong and does not hold in general. This section provides an overview of applications of CP with hierarchical, matrix/tensor, and graph data and additional assumptions required for the exchangeability to hold.

\subsubsection{Data with hierarchical structure} 
When data are generated via a hierarchical sampling procedure, it creates a simple form of heterogeneity within data. Consider the setting with $K$ many groups of observations and
$\Pi_k \sim P_{\Pi}, \quad N_k\mid \Pi_k \sim P_{N \mid \Pi}, \text{~independently~for~} k \in [K],$
where $N_k$ is the observed sample size for the $k$th group. For each group $k$, we have observations $\{(X_{k,i},Y_{k,i})\}_{i=1}^{N_k}$. As mentioned in \cite{dunn2023distribution}, there are two tasks of interest: (1) constructing a prediction set for the new sample $(X_{K+1,1},Y_{K+1,1})$ drawn from a new group $\Pi_{K+1} \sim P_{\Pi}$; (2) constructing a prediction set for the new sample $(X_{\mathrm{test}},Y_{\mathrm{test}})$ from one of existing $K$ groups.
Let $\tilde Z_k = (Z_{k,i})_{i=1}^{N_k} = ((X_{k,i},Y_{k,i}))_{i=1}^{N_k}$ for $k\in[K+1]$. 
As shown in \cite{dunn2023distribution}, in this setting, the hierarchical exchangeability condition is assumed as an extension of the generic exchangeability assumption, where not only are observations within each group exchangeable, $K$ groups are also assumed to be exchangeable.
% Then, we assume that for any permutation $\sigma$,
% \[
% P(\tilde Z_1, \tilde Z_2, \cdots, \tilde Z_{K+1}) \overset{d}{=} P(\tilde Z_{\sigma(1)}, \tilde Z_{\sigma(2)}, \cdots, \tilde Z_{\sigma(K+1)})
% \]
% and moreover, 
% \[
% P(\tilde Z_1, \cdots, (Z_{k,1},\cdots,Z_{k,m}), \cdots, \tilde Z_{K+1}) \overset{d}{=} P(\tilde Z_1, \cdots, (Z_{k,\sigma(1)},\cdots,Z_{k,\sigma(m)}), \cdots, \tilde Z_{K+1}) \mid N_k=m.
% \]

For task (1), four methods are proposed in \cite{dunn2023distribution}. The first approach, \emph{pooled CDF}, utilizes the estimated CDF from each group to construct asymptotically valid prediction sets.
The second approach, \emph{double conformal}, is similar to the quantile-of-quantile approach and achieves finite-sample validity under the hierarchical exchangeability assumption but tends to be conservative in practice.
The third approach, \emph{subsampling once}, samples one sample from each group and constructs the prediction set following the generic CP procedure without the requirement of hierarchical exchangeability. The shortcoming is that this approach does not utilize all the available data and may exhibit high variance. In light of this, \emph{repeated subsampling} repeats the procedure $B$ times to derandomize the subsampling procedure via p-values aggregation.

Another work \citep{bhattacharyya2024group} considers the setting with training and test data drawn from different multinomial distributions,
% \[
% X^0 \sim {\rm Multinomial}(p_1,\cdots,p_K),\;\;(X^1,Y) \mid (X^0=k) \sim P_k
% \]
% while the test sample is drawn from a different distribution
% \[
% X_{\rm test}^0 \sim {\rm Multinomial}(q_1,\cdots,q_K),\;\;(X_{\rm test}^1,Y_{\rm test}) \mid (X_{\rm test}^0=k) \sim P_k.
% \]
and to account for the heterogeneity in the sampling probabilities, weighted CP is adopted to adapt the test sample to ``similar'' observations.

\subsubsection{Matrix/tensor data}
An important line of research in unsupervised learning focuses on matrix estimation/prediction, e.g. forecasting in panel data with spatio-temporal correlations. In this setting, we have an underlying matrix $M \in \RR^{d_1 \times d_2}$ of interest, and each entry $(i,j)$ is missing following a certain pattern. With the random set of observed items $\calS \subseteq [d_1] \times [d_2]$, the task is to construct intervals $\{C(i^*,j^*): (i^*,j^*) \in \calS^c\}$ such that $M_{i^*j^*} \in C(i^*,j^*)$ with a high probability. 

In \cite{gui2023conformalized}, it is assumed that the matrix $M$ is deterministic and each entry $(i,j)$ is observed with probability $p_{ij} \in (0,1)$ independently and weighted exchangeability of indices in $\calS_{\cal} \cup \{(i^*,j^*)\}$ conditioning on $\calS \setminus \calS_{\cal}$ is established for $(i^*,j^*)$ uniformly drawn from $\calS^c$, which enables the splitting approach that uses $\calS \setminus \calS_{\cal}$ for estimating quantities of interest. More concretely, with $\hat M$, $\hat p$, and the local uncertainty score $\hat s$ estimated with $M_{\calS \setminus \calS_{\cal}}$, we have $V_{ij} = (|M_{ij} - \hat M_{ij}|)/\hat s_{ij}$ for all $(i,j) \in \calS_{\cal}$.
Then, by weighted CP with $\hat w_{ij}$ as the normalized odds ratio, the prediction interval for each $(i^*,j^*)$ can be constructed by $C(i^*,j^*) = \hat M_{i^*j^*} \pm \hat s_{i^*j^*} \hat Q$ where 
\begin{equation}
\hat Q = {\rm Quantile}_{1-\alpha}\left(\sum_{(i,j) \in \calS_{\cal}} \hat w_{ij} \delta_{V_{ij}} + \hat w_{i^*j^*}\delta_{+\infty}\right).
\end{equation}
This approach archives the validity in $\calS^c$ on average such that
\begin{equation}
\frac{1}{|\calS^c|} \sum_{(i^*,j^*) \in \calS^c} \ind\left\{M_{i^*j^*} \in C(i^*,j^*)\right\} \geq 1-\alpha-\Delta,
\end{equation}
where $\Delta$ measures the estimation accuracy of $\hat p$.

In comparison, \cite{shao2023distribution} considers the setting with a random matrix and fixed missing patterns. Valid intervals are constructed under the row/column exchangeability assumption on the random matrix. Since the coverage guarantee in \cite{gui2023conformalized} is on average for $\calS^c$, a recent work \cite{liang2024structured} considers a different criterion of coverage, \emph{simutaneous coverage}, which constructs a joint confidence region for $M$. \cite{sun2024conformalized} further extends the problem to the scenarios with a noisy and partially observed tensor.

\subsubsection{Data with graph (tree) structure}
Graph data is widely used across various fields, such as recommendation systems \cite{he2020lightgcn} and drug discovery \cite{gaudelet2021utilizing}. CP for graph data presents both unique challenges and opportunities. One key challenge arises from the reliance of standard CP on the exchangeability assumption, which may not hold for graph data due to its inherent dependencies. Several studies \cite{huang2024uncertainty,clarkson2023distribution,zargarbashi2023conformal} have explored the conditions under which the exchangeability assumption applies to graph data. A common consensus is that in inductive settings, where test data nodes/links are not present in the calibration data, the exchangeability assumption typically does not hold. However, in transductive settings, where all nodes from both the training and testing datasets are included in the graph, the assumption is more likely to be valid. Despite this challenge, the unique structure of graph data also presents opportunities to enhance the performance of CP methods, offering untapped potential for improvement.

% However, inaccuracies in these scenarios can lead to unreliable predictions, which may have serious consequences. Therefore, uncertainty quantification (UQ) methods, including conformal prediction (CP), are essential for ensuring the reliability of predictions in these scenarios. Some existing graph-based CP methods focus on the inductive setting, where test data nodes are not present in the training data, challenging the exchangeability assumption of CP. Other approaches seek to optimize prediction sets using graph structures to address CP inefficiencies. These methods are typically applied in the transductive setting, where all nodes from both training and testing datasets are included in the graph, thus maintaining exchangeability.

\textbf{Inductive setting}
In the inductive setting, models are trained on one graph (or a subset of nodes and edges) and then applied to unseen data, such as new nodes or entirely new graphs, during inference. As a result, the exchangeability assumption does not hold, causing CP to fail to provide coverage guarantees. To address this challenge, \citet{clarkson2023distribution} propose the Neighbourhood Adaptive Prediction Set. By leveraging the adjacency matrix, they generate prediction sets for test data, using the neighborhoods of test nodes as calibration data and introduce a weighting scheme based on the path length between nodes to adjust the importance of nodes in calibration, inspired by the homophily principle. In a different approach, \citet{zargarbashi2023conformal} note that the previous method may not work well with realistic sparse datasets, such as social network interactions. To tackle this issue, they suggest recalculating the nonconformity score and quantile each time new nodes are added to the graph. For edge classification, the same strategy is employed, but with a weighted nonconformity score defined as \( 1/\text{degree}(v) \), where \( v \) represents a node, to account for covariate shift.
In \citet{calissano2024conformal}, they proposed a CP method applicable to both labeled and unlabeled graphs. For unlabeled graphs, they embed the prediction set into a quotient space, where isomorphic graphs are grouped into equivalence classes. As a result, this approach effectively handles graph structures without reliance on specific node labels.

\textbf{Transductive setting}
In this setting, research has shown that exchangeability holds for both node and link prediction tasks \cite{huang2024uncertainty, zhao2024CLR}. As a result, most works focus on improving efficiency. In \cite{zargarbashi2023conformalp}, the authors update the nonconformity score by incorporating the nonconformity scores of neighboring nodes, effectively leveraging the graph structure to improve prediction efficiency. Specifically, they propose a diffused score as the nonconformity score:
\begin{equation}
       \hat{V}(v,y) = (1 - \lambda) V(x,y) + \frac{\lambda}{|\mathcal{N}_x|} \sum_{u \in \mathcal{N}_x} V(u,y) ,
\end{equation}
where \(\mathcal{N}_x\) is the set of neighboring nodes of node \(v\), and \(\lambda\) is a hyperparameter. \citet{huang2024uncertainty} further propose a more general method, highlighting that current approaches mainly focus on designing nonconformity scores while overlooking the direct training for efficiency. Based on the empirical finding that nodes linked together often produce similar sizes of prediction sets, they employ a second GNN model to help aggregate neighborhood information and introduce an inefficiency loss, defined as the size of the prediction set.

%%%%%%%%%%%%%%%%%%%%%%%%%
In addition to the aforementioned node classification tasks, there are some works that focus on applications of CP on link prediction. For example, extending previous exchangeability results in \cite{huang2024uncertainty}, \citet{zhao2024CLR} prove that the assumption also holds for link prediction tasks in GNNs, and introduce conformalized link prediction. The key finding is that graphs with node degrees following the power law distribution more closely tend to display smaller average prediction intervals, indicating higher efficiency. They then propose a sampling-based CP method to align the node degree distribution of the graph with a power-law distribution to enhance the efficiency of the CP process.
Another study \cite{marandon2024conformal} develops a conformal link prediction method that provides a set of true edges while controlling the False Discovery Rate (FDR). The motivation is that most methods consider link prediction as a ranking problem, which is satisfying when the application constrains the number of pairs of nodes to be declared as true edges to be fixed. Alternatively, other practices may require a classification of missing links into true and false edges with a control of false positives, e.g., reconstruction of biological networks. They collect a set of observed false links (those confirmed to not exist) and calculate p-values on these. Then, they apply the Benjamini-Hochberg procedure to the conformal p-values to establish a threshold for determining whether a link is a true connection. In a follow-up study, \citet{blanchard2024fdr} validate the effectiveness of this approach and extend it to include False Discovery Proportion bounds.

Another area of link prediction is recommendation systems, which aim to predict interactions between users and items. For example, \citet{angelopoulos2023recommendation} focus on controlling FDR for evaluating variable-sized sets in drug recommendation. To achieve this, they propose a method that calibrates learning-to-rank (L2R) models to maintain FDR control. Additionally, in \cite{wang2024confidence}, the authors introduce two novel CP-based losses: CP Set Size (CPS) and CP Set Distance (CPD). CPS is designed to improve the efficiency of CP, while CPD minimizes the cosine similarity distance between the top-K closest items in the CP set and the ground truth. This enhances recommendation precision by ensuring that items are not only selected with confidence (as with CPS) but also closely match user preferences. They train the model using joint learning to achieve these goals.

\textbf{Discussion} 
CP for graphs is a relatively underexplored area, and several challenges remain. First, the graph structure itself can be more thoroughly leveraged. Current research primarily focuses on neighborhood information, but other relationships--such as nodes with similar degrees or shared topological properties--could also be incorporated into CP. Second, there is limited research on applying CP to weighted graphs, such as transportation networks, or other graph variants, which presents a significant gap in the literature. Additionally, optimizing the efficiency of CP in settings beyond the current focus remains an open area for exploration.
\section{Static Data---Unstructured Data}
% \begin{figure}%[!ht]
%   \centering
%   \includegraphics[width=0.8\linewidth]{samples/nlp.pdf}
%   \caption{\lu{A taxonomy of CP methods for text data.}}
%   \label{fig:nlp}
% \end{figure}
\begin{wrapfigure}{l}{0.7\textwidth}
\vspace{-4mm}
    \centering
    \includegraphics[width=\linewidth]{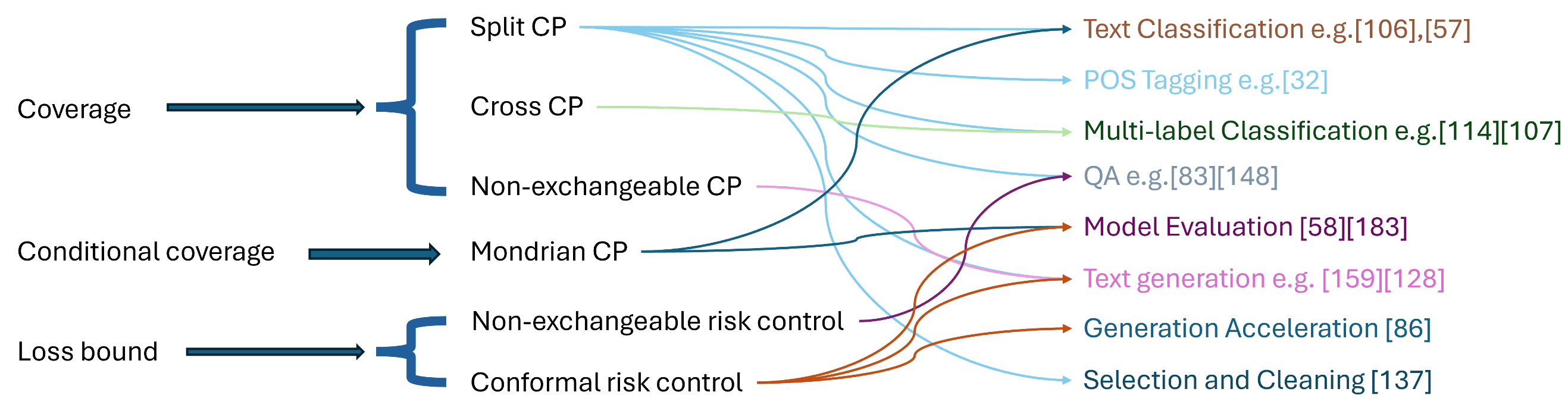}
    \caption{A taxonomy of CP methods for text data. Loss bound refers to a formal guarantee on the maximum expected loss of a prediction set generated by a conformal method.}
    \label{fig:<label>}
    \vspace{-4mm}
\end{wrapfigure}

\subsection{Text Data}
The rapid expansion of large language models (LLMs) and NLP applications has highlighted the critical need for UQ to mitigate risks such as hallucinations and to improve decision-making reliability in high-stakes scenarios. Text and LLMs possess unique characteristics that engender new challenges in CP due to their complex nature and unbounded output space. These LLMs are often trained on vast, inaccessible datasets and parameters, making re-training or even fine-tuning nearly impossible. CP addresses uncertainty without oversimplifying the complexity of LLMs and avoids the need for retraining and can be applied to any LLM. Primarily used in classification and regression tasks, harnessing the full potential of CP for text data and generative models such as LLMs is not trivial. In this section, we will explore CP in unstructured text data. For a more comprehensive review of CP for NLP, please refer to the survey \citep{campos2024conformal}.
%Add figure from jung2024trust
\vspace{-2mm}
\subsubsection{Text Classification}
Text classification is often applied in high-stakes areas such as medical diagnosis, security, and legal document review. 
% Despite achieving high accuracy, these applications can still have significant costs or consequences if not backed by robust statistical guarantees.
For example, \citet{maltoudoglou2020bert} build CP \cite{papadopoulos2002qualified} upon the transformer method for sentiment analysis to address the problem of heavy computation of CP. Another work \cite{giovannotti2022calibration} similarly applied inductive Venn–ABERS predictor to transformers and constructed prediction sets based on conformal p-values calculated from estimated probabilities.                
Recent work focuses on the conditional coverage of CP for text classification, which focuses on the coverage for certain subpopulations (e.g., conditional on labels) and is particularly essential for unbalanced datasets and respective tasks. \citet{giovannotti2021transformer} compare several transformer-based conformal predictors and found that Mondrian CP demonstrated the smallest theoretically expected efficiency drop for conditional coverage of CP.

Another thread of work focuses on multi-label text classification, where each instance may have multiple labels, requiring CP to output a combination of labels. For example, \cite{paisios2019deep} applies the power-set approach \cite{papadopoulos2014cross} to multi-label classification, which treats each label set as a candidate classification and is assigned a $p$-value. They also propose to only use the maximum observed label cardinality of the corpus to reduce the combination of labels. In the study \cite{maltoudoglou2022well}, they extend the Label Powerset (LP) Inductive (i.e., Split) CP by reducing computational complexity, specifically by eliminating a significant number of label sets that are likely to have $p$-values below the specified miscoverate rate. \cite{fisch2022conformal} proposes a modified CP algorithm that produces multi-label prediction sets with a strict limit on false positives. By using a greedy algorithm based on the nonconformity scores, it optimizes true positive rates while controlling false positives below a user-specified level.

\subsubsection{CP for Natural Language Understanding (NLU)}
%  NLU is becoming increasingly pivotal
% in various commercial settings. However, errors made by these systems can potentially result in
% significant consequences. To mitigate the impact of such mispredictions, NLU systems must assess
% their own uncertainty. 
NLU focuses on enabling machines to comprehend and interpret the structure and semantics of human language, it is becoming increasingly pivotal in various commercial settings. NLU involves analyzing text to extract meaning, intent, and context, handling ambiguity, and recognizing entities or relationships.
% Patrizio Giovannotti et al. \cite{giovannotti2022calibration} propose an inductive Venn–ABERS predictor (IVAP) applied to natural language understanding models. It functions to calibrate model outputs by ensuring that predicted probabilities are well-calibrated, reducing the uncertainty in binary classification tasks. This is achieved by training a transformer model and applying isotonic regression to refine the probability estimates, resulting in more accurate and reliable probabilistic predictions.
% \\\textbf{QA.} 

One application of NLU is \emph{question answering (QA)} tasks, which can be categorized into Multiple Choice Question Answering (MCQA) and open-domain QA. For the MCQA task, a common approach is to simplify it into a classification task for the given choices. Current approaches to applying CP to this task primarily rely on the standard split CP and developing various nonconformity scores for different needs. For example, one line of research~\citep{kumar2023conformal, ye2024benchmarking} involves appending the input with ``The correct answer is the option:'', prompting the LLMs to predict the next word to acquire the logits of the option. Subsequently, a softmax operation is applied to these logits and then used for CP algorithm. In open-domain QA, an LLM receives a question and is asked to generate an answer from a vast, unbounded space based on its comprehension of the question. In a black-box setting, when sampled answers as the LLM output are available, CP can be used to construct a reliable set of potential answers. For example, \citet{su2024api} employs sampling to assess the model's uncertainty and calculates nonconformity scores based on frequency to construct prediction sets. They also distinguish the results using normalized entropy and semantic similarity as an extension. 
Another application of NLU in open-domain QA is to detect LLM hallucinations and verify the facts. Li et al. \cite{li2024traq} construct a prediction set by sequentially building the retriever and response set with a coverage guarantee and then combining them into an aggregated set. Then it employs Bayesian optimization to enhance efficiency by finding the best hyper-parameters. An earlier work \cite{fischefficient} first expands the CP correctness
criterion to allow for additional, inferred “admissible” answers, which can substantially reduce the size of the predicted set while still providing valid performance guarantees. It then amortizes costs by conformalizing prediction cascades, in
which it aggressively prunes implausible labels early on by using progressively stronger classifiers—again, while still providing valid performance guarantees.

Another three common NLU tasks where CP has been studied are \emph{machine translation evaluation}, text infilling, and \textit{part-of-speech (POS) tagging}. Machine translation evaluation involves assessing the quality of generations in a target language based on the input provided in a source language. 
Some studies apply the CP algorithm to machine translation evaluation to address uncertainty \cite{zerva2023conformalizing, giovannotti2023evaluating}. For example, \citet{zerva2023conformalizing} show that split CP adjusts confidence intervals to achieve the desired coverage and identify biases in these intervals based on translation language pairs and translation quality. Using conditional CP, they create calibration subsets for each data subgroup, ensuring equalized coverage. In the POS tagging and text infilling, \citet{dey2022conformal} propose CP-enhanced BERT and BiLSTM algorithms to ensure their statistical reliability. The non-conformity scores are designed based on multi-label classification tasks. 

\subsubsection{CP for Natural Language Generation (NLG)}

NLG is the process of transforming structured information or text into natural language output. It focuses on producing coherent, contextually appropriate language, such as generating summaries or translations.

In the field of NLP, transformer-based generative LLMs play a significant role \citep{vaswani2017attention}. These models generate text by predicting the next word based on previous inputs.
% creating a probability distribution and selecting the next word through methods like greedy search or sampling. 
When applying CP to text generation tasks, these tasks are categorized by the length of the generated content, distinguishing between next-word prediction and long sentence generation tasks.
We note that different from MCQA, in the next-word prediction task, the label set encompasses the entire vocabulary, from which CP generates a prediction set containing potential next words. 

% The difference between MCQA and the next-word generation lies in the fact that, in the next-word generation, the number of possible next choices corresponds to the entire vocabulary size. 
To assess whether a top-$p$ sampling set is indeed aligned with its probabilistic meaning in various
linguistic contexts, \citet{ravfogel2023conformal} leverage CP and customize the Adaptive Prediction
Sets (APS) score \cite{romano2020classification}, which aims to provide prediction sets that adaptively include all possible labels that meet a certain confidence level, ensuring that the true label is contained within the prediction set with a high probability. 
Another challenge in next-word prediction is that the complex intrinsic relationships within the data may violate the exchangeability assumption, leading to non-i.i.d. test and calibration data. To address this, Dennis Ulmer et al. \citep{ulmer2024non} combine APS scores with non-exchangeable CP for next-word prediction.

Another increasingly popular NLG task is \emph{code generation}, where LLMs are tasked with producing a piece of code based on input. One challenge lies in structured prediction problems, where the space of labels is exponential in size. To address this issue, Adam Khakhar et al.~\cite{khakhar2023pac} propose generating code based on the abstract syntax tree structure and partial programs. They utilize partial programs for prediction~\cite{solar2008program} and establish a pre-selected 1D search space over the prediction sets by mapping features to labels in order to construct Probably Approximately Correct (PAC) prediction sets. 
\emph{Retrieval Augmented Generation (RAG)} combines retrieval-based methods with LLMs for text generation. Integrating CP in its unique generation process is challenging.
Kang et al. \cite{kangc} propose to apply conformal risk control with RAG, in which they show that using RAG can lead to lower risk than normal LLM with or without covariate shift.
Mohri et al. \cite{mohrilanguage} address the problem of factual correctness in LLM outputs. During calibration, each output is broken into subclaims, scored by GPT-4 for confidence, and filtered using a threshold based on a preset confidence rate $\alpha$. In the prediction stage, subclaims with confidence scores above this threshold are seen as correct factual and concatenated to form the final output. 
% \textbf{Conformal Alignment} 
% \lu{we need to define what alignment is. There are many different definitions} 
% In radiology report generation, reports generated by a vision-language model must align with human evaluations before their use in medical decision-making. 

Another line of research adopts CP to control the quality of LLM generations. To deploy the outputs of foundation models reliably, \citet{gui2024conformal} present conformal alignment, which aims to provide guarantees that a prescribed fraction of selected model outputs are trustworthy and meet the alignment criterion. This is achieved by training an alignment predictor on reference data and selecting outputs whose predicted alignment scores exceed a calibrated threshold, thus controlling the false discovery rate of untrustworthy outputs.
Victor et al. \cite{quachconformal} introduce a conformal sampling framework for LLMs to generate diagnoses based on input content. Its goal is to generate prediction sets with provable guarantees of containing at least one acceptable response, while filtering out low-quality or redundant outputs. This is achieved through a calibrated sampling process that stops when confidence criteria are met, using rejection rules to eliminate undesirable samples, ensuring efficient and reliable output.
\vspace{-4mm}
\subsubsection{Inference-time Efficiency} 
Besides generating prediction sets, CP can also improve inference-time efficiency by reducing the need for multiple samples or complex generation and leveraging efficient calibration strategies. Some works focus on achieving early stopping using CP, enabling the model to produce reliable predictions with reduced inference time. In \cite{laufer-goldshtein2023efficiently}, the authors propose training an LLM using minimal computational resources while maintaining accuracy. By leveraging the Learn then Test (LTT) framework, they identify optimal risk-controlling hyperparameter configurations. Their key contribution, Pareto Testing, further improves efficiency by strategically guiding the number and order of tested configurations when searching for valid settings.
Schuster et al. \cite{schuster2022confident} propose a novel method called Confident Adaptive Language Modeling (CALM), which allows the transformer to exit early at a specific layer of the decoder, thereby significantly reducing inference time. This method employs various confidence measures to determine when to exit early, such as softmax response, hidden-state saturation, and a trained early exit classifier, then they uses a calibration dataset to threshold confidence scores by grid search.
% This involves performing a grid search over possible $\lambda$ values to find the optimal threshold that balances computational efficiency and model performance. The calibration ensures that the early-exit decisions do not degrade the overall quality of the generated text. 
\\
\noindent\textbf{Discussion}
Although CP has recently achieved success in NLP, several questions remain to be addressed in the future: (1) Computational efficiency. The use of CP can be computationally expensive, especially with large datasets or packages like FAISS\cite{johnson2019billion}. Optimizing CP for efficiency is essential.
(2) Human and LLMs Interaction. As human-LLM interactions generate continuous new data, further research is needed to adapt CP to effectively and efficiently handle these unbounded output space and dynamic environments.
(3) Ethical Concerns. CP's ability to produce multiple predictions increases the risk of generating unethical outputs, particularly in LLMs, and addressing these concerns remains an open challenge.

% 3. Fairness: Fairness in NLP is a significant concern. Models trained in a specific domain may not generalize well to other domains, leading to data bias. Addressing this bias and improving the fairness of CP in NLP is another important direction for future research.
\subsection{Image Data}
%% specialty of CP in image
Deep learning with images can achieve high predictive accuracy, but quantifying uncertainty remains a significant challenge, hindering its deployment in critical settings. Applying CP for image data involves navigating challenges related to high dimensionality and feature complexity. The following sections will introduce the development of CP in various computer vision (CV) tasks.
\subsubsection{Image classification} 
% In the CV classification task, most related papers focus on proposing a new CP framework. 
Some applications of CP for image classification are built upon the framework of canonical classification tasks, but further exploit the specific characteristics of image data across various domains to design powerful nonconformity score functions.
% For the classification task, given an image $x$, we want to find a function $f(x)->y, y\in \mathbb{Z}^+$, which can judge the class of give image.
% For CP, For each image $x$ we build a interval in the following formula:
% \begin{align}
%     \mathcal{T}\left(x\right) = \left[c_1, c_2, c_3,...,c_n\right]
% \end{align}
% Where $\hat{l}$ and $\hat{u}$ is the lower and upper bound of the predicted interval.
% To guarantee the coverage of the predicted interval $\mathcal{T}$, the user choose a risk level $\alpha \in \left(0, 1\right)$ and an error level $\sigma \in \left(0, 1\right)$, for CP, we will make sure:
% \begin{align}
%     \mathbb{E}\left[\frac{1}{|D_{test}|}\left|\left\{i: y_{i} \in \mathcal{T}(x)_{i}|i\in D_{test}\right\}\right|\right] \geq 1-\alpha
% \end{align}
% Where $D_{test}$ is the test dataset.
%%%%%%%%%%%%%%%%%%%%full CP%%%%%%%%%%%%%%%%%
%一段
% \textbf{Conformal Triage for Medical Imaging AI Deployment} 
% \z{ This seems like an application of CP in medical imaging. Are there any differences in applying CP here? I guess the design of parameters here is related to controlling the false positive rate and false negative rate. If so, this appears to be a unique objective in medical image classification that hasn't been discussed in the previous section.} 
For example, in \cite{angelopoulos2024conformalimage}, they aim to ensure the reliability of AI models in diverse clinical environments by introducing the conformal triage algorithm, which outputs labels \{true, false, and unknown\} to aid clinical decision-making.
% In their method, they aim to predict a binary Y from feature X, they propose to find a $\lambda_{PPV}$ and a $\lambda_{NPV}$, given these two parameters, the prediction is output by:
% \begin{align}
%     \hat{Y}(X) = 
% \begin{cases} 
% + & \text{if } f(X) \geq \lambda_{PPV} \\
% - & \text{if } f(X) \leq \lambda_{NPV} \\
% ? & \text{else}
% \end{cases}
% \end{align}
% To find these two paramters, they use a calibrate data, the $\lambda_{PPV}$ is a mininum $\lambda_{PPV}$ which guarantees the upper end of the binomial confidence interval for (true positive)/(predicted true samples) achieves a presetting coverage, the $\lambda_{NPV}$ is designed  is constructed analogously.
% In their experiment, the algorithm was tested on head CT scans from Massachusetts General Hospital using a pre-existing model for detecting intracranial hemorrhage. The results shows:
% The high-risk group achieved the desired PPV and the low-risk group achieved the desired NPV, demonstrating the reliability of the conformal triage algorit conformal triage algorithmhm in clinical settings.
%%%%%%%%%%%%%%%$ CP from other view%%%%%%%%%%%%%
Another line of research utilizes learned deep (underlying) features of image data in constructing more efficient prediction sets.
Teng et al.~\cite{teng2023predictive} extend the application of CP to semantic feature spaces, in which they leverage the difference between pre-trained and fine-tuned features to generate the prediction set via band estimation~\cite{xu2020automatic}.
% Teng et al.~\cite{teng2023predictive} extend the application of CP to semantic feature spaces, which provide richer information for exploiting semantic features. Furthermore, their approach can be seamlessly integrated as a plug-in component with any pre-trained model. In their method, they leverage the difference between pre-trained and fine-tuned features, along with non-conformity scores, and utilize band estimation~\cite{xu2020automatic} and band detection to generate the prediction set.
\citet{ghosh2023improving} further show that if learned image representation satisfies some mild conditions, their proposed NCP (neighborhood conformal prediction) approach will produce smaller prediction sets than traditional CP algorithms. They utilize the learned representation to identify the $K$-nearest neighbors of a given test input and assign important weights based on their distances to create adaptive prediction sets. 

\subsubsection{Image segmentation}
Given an image $x$ as the input, image segmentation aims to find a map $f: x \mapsto y$, where $y$ is the segmentation result as the output.
Here $x$ and $y$ have the same dimension and for each pixel $x_i$, the corresponding value $y_i$ represents the class for this pixel. As the softmax function has been shown to produce overconfident predictions with uncalibrated outputs, CP can be beneficial in calibrating the segmentation approaches. 
% In CP, the goal is to calibrate the data and determine a function that predicts a prediction set $[c^i_1, c^i_2,...,c^i_n]$for each \( y_i \).
% \textbf{Pixel-wise prediction sets.}
\cite{wieslander2020deep} combines deep learning with CP for tissue sub-region prediction with confidence. 
They define the nonconformity score based on the output softmax score for each pixel. They further use the conformal $p$-values to guide prediction: they first choose the class with the largest $p$-value as the predicted output. Then, confidence in the prediction is defined as the gap between the largest and the second largest $p$-values, with which the pixel with confidence lower than a given confidence level is excluded.
In addition, a recent work \cite{mossina2024conformal} utilizes the framework of conformal risk control, where the nested sets $\{\hat C_{\lambda}(\cdot)\}_{\lambda \in \Lambda}$ is defined by thresholding the output softmax scores by $\lambda$.
% \textbf{Calibration beyond pixels.}
The scarcity of images for pixel-level calibration makes it hard to obtain sufficient images for pixel-level calibration. \citet{brunekreef2024kandinsky} introduce the concept of Kandinsky calibration, which leverages the inherent spatial structure within natural image distributions, to cluster pixels. 
% In this method, they cluster pixel according to a defined nonconformity curve. Accordingly, at the cluster level, they aggregate all the non-conformity scores in this cluster to compute a cluster-specific nonconformity curve. 

\subsubsection{Image-to-image tasks} 
% \lu{which paper is it?}
% Conventional image-to-image methods lack statistical guarantees for generated results, limiting their application in high-stakes scenarios. Therefore, applying CP to image-to-image tasks is highly valuable. 
In image-to-image tasks, including inpainting, super-resolution, and colorization, the objective of CP is to construct a prediction interval around each pixel such that the true pixel value falls within this interval with a user-defined probability.

% In the image-to-image task, given an original image \( y \in \mathbb{R}^d \), we aim to find a function \( f(y) \rightarrow x \), where \( x \in \mathbb{R}^d \), which can transform the original image into the target image.
% For CP, we construct an interval for each pixel \( x \) using the following formula:
% \[
% \mathcal{T}(x_{mn}) = [\hat{l}_{mn}, \hat{u}_{mn}]
% \]
% where \( m \) and \( n \) denote the pixel's position in the image, and \( \hat{l} \) and \( \hat{u} \) are the lower and upper bounds of the predicted interval, respectively.
% To ensure the coverage of the predicted interval \( \mathcal{T} \), the user selects a risk level \( \alpha \in (0, 1) \) and an error level \( \sigma \in (0, 1) \). For CP, we ensure:
% \[
% \mathbb{E} \left[ \frac{1}{M N} \left| \left\{ (m, n) : y_{(m, n)} \in \mathcal{T}(x)_{(m, n)} \right\} \right| \right] \geq 1 - \alpha
% \]
% and for risk control, we ensure that the probability of the above condition is at least \( 1 - \sigma \).
% Principal UQ， Semantic uncertainty intervals for disentangled latent spaces 结合
Canonical approaches to this task define uncertainty regions based on probable values per pixel, while ignoring spatial correlations within the image, resulting in an exaggerated volume of uncertainty.
To address this, one study \cite{sankaranarayanan2022semantic} quantifies the uncertainty in a latent space, thus
taking spatial dependencies into account. Specifically, they aim to develop a method that can provide principled uncertainty intervals for semantic factors within a disentangled latent space of a generative model. This method has a fixed decoder that can predict images based on latent code with conformal quantile regression and uses a decoder to generate restored images from the prediction set. However, by relying on a non-linear, non-invertible and uncertainty-oblivious transformation, this method suffers from interpretability limitations. \citet{belhasin2023principal} improve this by using spatial relationships between pixels for UQ in inverse imaging problems. They employ a diffusion model with three phases: (1) approximation, where sampling generates principal components and weights using SVD; (2) calibration, where three variants of Principal UQ (PUQ) are proposed: exact PUQ, Dimension-Adaptive PUQ, and Reduced Dimension-Adaptive PUQ, with the latter two improving computational efficiency; and (3) using the previous result for prediction.

% the data set is $(X,Z)$, where $X$ is corrupted image and $Z$ is its corresponding latent code in a disentangled latent space, they use quantile regression to train a encoder for predicting latent codes, which output $q_{\alpha/2(X)}$ and $q_{1-\alpha/2(X)}$, and is calibrated by a calibration dataset to ensure they contain the true latent values with high probability. The $q_{cal, \alpha/2(X)}$ and $q_{cal,1-\alpha/2(X)}$ is then used to generate restored image by decoder $G$.
% \z{ I am having difficulty following the logic of this section. Could we reorganize the content with more informative titles?}
% \textbf{Confidence interval for diffusion model} To improve the sampling efficiency and reduce calculation resource of CP in diffusion model,  
% \textbf{Conformal Risk Control}

There are several works that focus on conformal risk control for image-to-image tasks. For example,  \cite{angelopoulos2022image} proposes a distribution-free pipeline for image-to-image regression based on risk control \cite{bates2021distribution}. \cite{kutiel2022s} proposes a method for conformal risk control based on masking, which guarantee the masked reconstructed image and masked true image have a distance under threshold in a given probability. \cite{sridhar2023diffusion} proposes a novel method for applying diffusion models to CP. 
Their approach involves generating multiple reconstructed variations for each original image and then determining the interval bounds by taking the upper and lower quantiles for each pixel across these variations. 
% They perform calibration using Hoeffding’s upper-confidence bound to obtain a scaling factor \( \lambda \). 
To deal with the problem that classical risk control can only be applied to scalar hyper-parameters, in the study \cite{teneggi2023trust}, they propose entrywise calibrated intervals and multi-dimensional risk control.  
% \z{ I suggest first illustrating the motivations or proposing the challenges they aim to solve here before presenting the methods.}
% First, the authors use CP to create entrywise calibrated intervals, providing high-probability coverage for future samples. They then extend this approach to a high-dimensional risk control procedure, called K-RCPS, which employs a convex optimization framework. The intervals are parametrized to minimize mean interval length while ensuring the intervals remain nested and the loss function is entrywise monotonically nonincreasing. The core optimization problem is solved using a surrogate loss function that simplifies the non-convex problem into a convex one. The K-RCPS algorithm then fine-tunes these intervals to guarantee risk control. The method is empirically validated on real-world datasets, demonstrating its effectiveness in producing tight, reliable uncertainty intervals for tasks such as denoising natural face images and CT scans, thereby enhancing the trustworthiness of diffusion models in critical applications.
\\
\noindent\textbf{Discussion} CP for image classification focuses on high efficiency, handling data imbalance or noise, and achieving the performance of full CP. These methods perform well because they address various challenges associated with CP. In image-to-image tasks, encoder-decoder structures like diffusion models are commonly used. However, there are still some problems that can be better addressed. For example, in image-to-image tasks, leveraging the relationship between the prediction sets of different pixels to improve predictions is an area for further exploration.
% Second, while full CP approaches have been developed, reducing their computational complexity remains an important research area.
\subsection{Heterogeneous Data}
% \lu{remove half page. No need for subsubsections, just use two paragraphs.}
Heterogeneous data consists of diverse types and formats, often originating from different sources. This diversity presents challenges for traditional CP methods, which are generally designed for more uniform data types. The variability in distributions, complexity in feature interactions, and the need to handle different data structures simultaneously make it difficult for traditional CP to provide accurate and reliable prediction intervals. Therefore, new CP methods are required to effectively manage heterogeneous data, ensuring that predictions remain reliable and maintain confidence guarantees across all types of data involved.
For heterogeneous data, the objective is to find a function that can map \(\mathcal{X} \rightarrow \mathcal{Y}\), where \(\mathcal{X} = \mathcal{X}_1 \cup \ldots \cup \mathcal{X}_K\).
Then in a standard CP method for heterogeneous data, for each sample \(x\), a prediction interval/set is generated by:
\vspace{0pt}
\begin{equation}
C(x) = [\hat{l}, \hat{u}] \text{ or } [c_1,c_2,..,c_n]
\end{equation}
\vspace{0pt}
where \(\hat{l}\) and \(\hat{u}\) are the lower and upper bounds of the predicted interval, respectively, $c_i$ is a label.
% To guarantee the coverage of the predicted interval \(\mathcal{T}\), the user chooses a risk level \(\alpha \in (0, 1)\) for CP. We ensure that:
% \begin{align}
% \mathbb{E}\left[\frac{1}{|D_{\text{test}}|} \left| \left\{ i : y_{i} \in \mathcal{T}(x_{i}) \mid i \in D_{\text{test}} \right\} \right|\right] \geq 1 - \alpha
% \end{align}
% where \(D_{\text{test}}\) is the test dataset.

One study \cite{pmlr-v242-tumu24a} works on multimodal data and highlights that previous methods for generating multi-modal prediction regions often result in overly complex solutions.
% , making them unsuitable for downstream tasks like planning and control. 
To address this, it proposes using shape templates—such as ellipsoids, convex hulls, and hyperrectangles—over the residuals of calibration data, which simplify the non-conformity score functions, leading to more efficient and accurate prediction regions. When there is noise in the multimodal data, \citet{zhan2023reliabilitybased} propose a reliability-based training data cleaning method using split CP, which leverages well-labeled data as a calibration set to detect and correct mislabeled data and outliers. In another study \cite{duttaestimating} for image caption, they introduce a novel heuristic for zero-shot CP that leverages CLIP-based foundation models \cite{radford2021learning} calibrated by web data. Some work develops CP for limited data scenarios with multi-modal training, e.g. zero-shot, one-shot, and few-shot learning. For example, \citet{fisch2021few} point out that when the dataset is small, the prediction generated by CP will be unusually large. They address it by casting CP as a meta-learning paradigm over exchangeable collections of auxiliary tasks. Another type of heterogeneous data is multiview data which refers to datasets where the same underlying object or phenomenon is described by multiple different sets of modes or ``views.'' 
Multi-View CP (MVCP) \cite{rivera2024conformalized} addresses the challenge that the naive method for late fusion fails to leverage the uncertainty information of individual predictors, leading to less efficient and adaptive prediction intervals. Specifically, they define the score function for the predictor in each mode and use quantiles along projection directions. The final prediction region is an intersection of half-planes, ensuring efficient, distribution-free coverage.

Another application of heterogeneous data is in federated or transfer learning, where the model processes multi-source inputs. In \cite{lu2023federated}, the authors highlight that data in federated learning is typically non-IID, as data from different sources often follow distinct distributions. This characteristic poses challenges to the non-exchangeability assumption of conformal prediction. To address this issue, they propose the concept of an $\epsilon$-approximate $\beta$-quantile, which allows the quantile to deviate slightly from the target value while maintaining similar coverage properties. Within this framework, non-conformity scores are computed locally for calibration before communication occurs. Further, in transfer learning, \citet{liu2024multi} study covariate shifts when data originates from multiple sources. The proposed data-adaptive strategy begins by calculating site-specific quantiles for each source. For the target distribution, it then computes a discrepancy measure between the target site's quantile and each source site's quantile. This discrepancy is used to assign weights and  minimize the discrepancy for each source site and the target site.
% Specifically, it utilizes the internet as a queryable source of calibration data. At test time, it generates prompts like ``an image of a <category>'' for zero-shot classification and retrieves class-specific calibration data from the web. Then it computes ``plausibility scores'' based on retrieved images and their contexts to refine the prediction process.
\\
\noindent\textbf{Discussion}
Existing CP methods for heterogeneous data are relatively limited, primarily focusing on multi-modal and multi-view data, as well as scenarios like covariate shift and limited data availability. Future research could explore leveraging contrastive learning frameworks, similar to CLIP for multi-modal image captioning, to enhance conditional probability CP efficiency. Another promising direction is addressing covariate shifts across different modalities in multi-modal tasks, which remains largely unexplored. Further, managing uncertainty in multi-modal outputs, which often arises from the complexity of such data, presents a compelling challenge that warrants further investigation. Finally, more efforts are needed to explore CP in the semantic space. 
% \section{Dynamic Data}
%%% discussion.
\section{Dynamice Data---Spatio-temporal Data}\label{sec:spatio-temp}
\begin{wrapfigure}{r}{0.38\textwidth}
    \centering
    \includegraphics[width=1\linewidth]{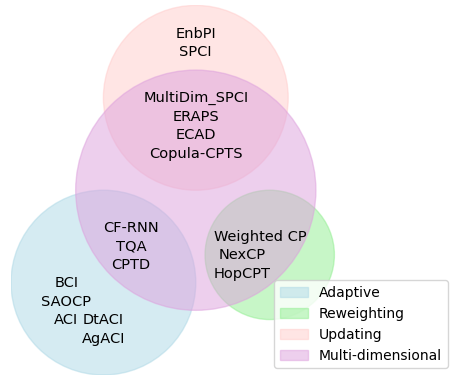}
    \caption{Venn Diagram for CP Methods in Spatio-Temporal Data.}
    % \vspace{-1mm}
    \label{fig:venn diagram}
\end{wrapfigure}

Spatio-temporal data captures phenomena that evolve over time and across locations, making it crucial in fields like meteorology, geosciences, urban planning, and public health. 
As prediction models for spatio-temporal data are often setting-specific, prediction sets instead of point predictions can be preferred in this scenario. 
Applying CP to this type of data helps manage risks associated with spatial and temporal variability by providing confidence measures for predictions. However, standard CP assumes data exchangeability, an assumption often violated by spatio-temporal data due to its structured dependencies and evolving nature. This can lead to misleading confidence measures if not properly addressed. This section discusses challenges and methodologies associated with applying CP to spatio-temporal data. We begin by discussing one-dimensional spatio-temporal data, which essentially constitutes time series data collected from a single geographic location. Then, we explore strategies for tackling multi-location (i.e., multi-dimensional) spatio-temporal data, where each entire series is treated as a single observation, exhibiting temporal dependencies both within and between the sequences. Therefore, the violation of exchangeability in multi-dimensional settings can be more severe and requires robust analytical approaches to ensure the accuracy and reliability of predictive models. In addition, we will cover the special case of streaming time series data, highlighting the need for new methods and identifying additional areas for the application of CP.  We summarize representative CP methods in this section in Table \ref{Table: spatio-temporal summary} and Fig. \ref{fig:venn diagram}.
% \lu{The visual comparison of these two settings is illustrated in Fig. \ref{fig:timeseries}.}
% arrow consistent
% \begin{figure}[htbp]
%     \centering
%     % 第一行图像
%     \begin{minipage}{.44\linewidth}
%         \centering
%         \includegraphics[width=\textwidth]{figure/one_dimensional.png}
        
%     \end{minipage}%
%     %\hfill%
%     \begin{minipage}{.56\linewidth}
%         \centering
%         \includegraphics[width=\textwidth]{figure/multi_dimensional.png}
%     \end{minipage}
%     \caption{\lu{CP for one-dimensional vs. multi-dimensional spatio-temporal data.}}
%     \label{fig:timeseries}
% \end{figure}
% Time-series Observation Paradigms. (Left) The dataset consists of a single time series, where each observation represents an individual time step with temporal dependencies \lu{(i.e., one-dimensional)}. (Right) The dataset consists of multiple time series, where each entire series is treated as a single observation, exhibiting temporal dependencies both within and between the sequences.
\begin{table}[htbp]
\setlength{\belowcaptionskip}{0pt} 
\centering
\resizebox{\linewidth}{!}{%
\begin{tabular}{ccccc}
\hline
Representative Work & Data & Distribution Assumption & Coverage Guarantee \\ \hline
NexCP \cite{barber2023conformal} & None & None & $\geq 1 - \alpha - \sum_{i=1}^{n} \tilde{w}_i \cdot {\text{d}_\text{TV}} \left( R(Z), R(Z^i) \right)$ \\ 
EnbPI \cite{xu2021conformal}  & single time series & strongly mixing errors & $\approx 1 - \alpha$ \\ 
SPCI \cite{xu2023sequential}& single time series & None & $\approx 1 - \alpha$ \\ 
ACI \cite{gibbs2021adaptive}& single time series & None (online) & Asymptotically $\approx 1 - \alpha$ \\ 
CF-RNN \cite{stankeviciute2021conformal} & independent time series & exchangeable time series & $1 - \alpha$ \\ 
Copula-RNN \cite{sun2023copula}  & independent time series & exchangeable time series & $1 - \alpha$ \\ 
MultiDimSPCI \cite{xu2024conformal}& dependent time series & non-exchangeable time series & $\approx 1 - \alpha$ \\ \hline
\end{tabular}%
}
\caption{A summary of representative CP methods for spatio-temporal data described in Sections \ref{sec:spatio-temp}.}
\label{Table: spatio-temporal summary}
\vspace{-7mm}
\end{table}

\subsection{One-Dimensional Spatio-temporal Data}
For one-dimensional spatio-temporal data, there are three primary lines of research:

\begin{itemize}
\item \textbf{Reweighting} adjusts the influence of historical samples based on their age or relevance, addressing the non-stationarity. By altering the weights assigned to each sample, 
sample reweighting helps to align the distribution of the weighted calibration data more closely with that of the test data, ultimately addressing the issue of distribution shift. 

\item \textbf{Updating non-conformity scores} updates non-conformity scores continuously as new data becomes available, ensuring that the model adapts to recent trends and patterns without being overly influenced by outdated information.

\item \textbf{Updating $\alpha$:} Instead of using a fixed miscoverate rate for all predictions, this method adjusts the miscoverate rate based on recent prediction errors and variability in the data. Such adaptivity allows for more flexible and context-sensitive prediction intervals, crucial for applications where sudden changes in volatility or trend are common.

\end{itemize}
% These approaches aim to provide robust confidence estimates, which are crucial for decision-making in sectors such as finance and meteorology, where accurate and trustworthy predictions are paramount. By modifying traditional CP techniques to better align with the properties of time series data, we enhance both the practicality and effectiveness of these models in real-world scenarios.

%\textcolor{blue}{introduce how to categorize these methods}
%\subsection{Adjust $\alpha$}
%\subsubsection{Update Non-Conformity Score}
%Extending CP to time-series data considers weighing the past non-conformity score non-equally so that scores more similar to the present are given high weights\cite{xu2024conformal}.
%\textcolor{blue}{introduce the motivation and structure of this section}
\subsubsection{Reweighting} %Reweighting is an adaptation within the CP framework, specifically designed to handle settings where the conventional assumption of exchangeability is violated. This includes scenarios such as label shift \cite{podkopaev2021distribution}, causal inference setups \cite{lei2021conformal}, survival analysis 
 %\cite{candes2023conformalized}, and other non-exchangeable datasets. Central to this method is the assignment of weights to each sample based on its relevance or similarity to the point for which a prediction is being made. While reweighting modifies the conventional CP procedure by incorporating these weights into the calculation of nonconformity scores, thereby accommodating the specific characteristics of various datasets, it is particularly well-suited to time series data. Hence, we discuss the details of this framework here, emphasizing its effective application in reflecting the temporal structure and dynamics typical of time series \cite{tibshirani2019conformal}. Initially, we will explore the application of reweighting to address covariate shift and distribution drift, employing predetermined, non-adaptive weights based on prior knowledge. Building on this foundation, we will then discuss the implementation of data-adaptive weights tailored for time-series data, leveraging ongoing insights to refine our predictive models.
% Reweighting is a general framework for non-exchangeable settings, which assigns relevance-based weights to samples such that the reweighted data distribution is closer to some ``target'' distribution, making it particularly effective for time series data by accommodating temporal structure and dynamics. 
Reweighting is one of the most common frameworks for non-exchangeable settings, including time-series data. As we show in Section 2.3, weighted CP can compute prediction intervals for problems with covariate shifts between training and test data \cite{tibshirani2019conformal}. Below, we discuss several reweighting approaches that can be applied to one-dimensional time series data.

\textbf{Pre-defined weights} 
To relax the covariate shift assumption and address the problem that the high-dimensional likelihood ratio is known beforehand or well approximated for correct coverage in weighted CP, \citet{barber2023conformal} extended the weighted method in \cite{tibshirani2019conformal} to the scenario with arbitrary pre-defined weights, showing that the coverage gap can be bounded by the total variation between the calibration and test data distributions. The weights $\{w_1, \ldots, w_t \in [0,1]\}$ for non-conformity scores are normalized as follows:
\begin{equation}
\tilde{w}_i = \frac{w_i}{w_1 + \ldots + w_t + 1}, \quad i = 1, \ldots, t, \quad \text{and} \quad \tilde{w}_{t+1} = \frac{1}{w_1 + \ldots + w_t + 1}.
\end{equation}
The weighted nonconformity scores for split and full CP are then:
\begin{equation}
\sum_{i=1}^{t} \tilde{w}_i \cdot \delta_{V_i} + \tilde{w}_{t+1} \cdot \delta_{\infty}
\quad \text{and} \quad
\sum_{i=1}^{t+1} \tilde{w}_i \cdot \delta_{V_i}^{y}, \text{ respectively.}
\end{equation}
 Here $V_i = |Y_i - \hat{f}(X_i)|$ is the non-conformity score for the trained model $\hat{f}$. Let $\text{d}_{\text{TV}}$ denote the total variation distance between distributions, theoretically, \cite{barber2023conformal} provided the lower bounds for coverage in nonexchangeable CP:
\begin{align}
\mathbb{P} \left\{ Y_{n+1} \in \hat{C}_n(X_{n+1}) \right\} \geq 1 - \alpha - \sum_{i=1}^{n} \tilde{w}_i \cdot {\text{d}_\text{TV}} \left( R(Z), R(Z^i) \right).
\end{align}

%From this result, we see that if \(\tilde{w}_{n+1} = \frac{1}{w_1+\ldots+w_{n+1}}\) is small (which corresponds to the effective sample size of our weighted method being large), then mild violations of exchangeability can only lead to mild undercoverage or to mild overcoverage.
This approach can be applied to non-symmetric algorithms like weighted regression and autoregressive models, which are common in time series forecasting. %The approach involves defining a non-symmetric algorithm $\mathcal{A}$ that maps sequences of tagged samples $(X_i, Y_i, t_i) \in \mathcal{X} \times \mathbb{R} \times \mathcal{T}$ to a regression function $\hat{\mu}$. Before fitting the model, tags of two samples are randomly swapped based on a multinomial distribution with weights $\tilde{w}_i$. The algorithm $\mathcal{A}$ is then applied to the permuted data, including a hypothesized test point $y$. Residuals $R^{y,K}_i$ are computed from the fitted model. Finally, the prediction set $\hat{C}_n(X_{n+1})$ is constructed by including all $y$ such that the residual $R^{y,K}_{n+1}$ falls below a quantile threshold $Q_{1-\alpha}$ of the weighted empirical distribution of residuals:
%\[
%\hat{C}_n(X_{n+1}) = \left\{ y : R^{y,K}_{n+1} < Q_{1-\alpha} \left( \sum_{i=1}^{n+1} \tilde{w}_i \cdot \delta_{R^{y,K}_i} \right) \right\}.
%\]
Building upon \cite{barber2023conformal}, subsequent research broadened the framework's applications. For example, \citet{schlembach2022conformal} expanded the methodology for multi-step-ahead time series forecasting to the multivariate setting and conformal risk control measures were incorporated into one-dimensional spatio-temporal data \cite{farinhas2023non}.

%The nonexchangeable CP framework enhances robustness against distribution drift by employing weighted quantiles and a novel randomization technique accommodating algorithms that handle samples asymmetrically. Its robustness is mathematically guaranteed, maintaining coverage reliability in the presence of distribution drift and other complexities typical in real-world datasets. This adaptability is particularly beneficial for time series data analysis.
%In advancing CP for time series, a pivotal consideration is the optimal selection of weights, a concept driven by nonexchangeable CP. Barber et al.\cite{barber2023conformal} proposed using exponentially decayed weights which, being independent of data, place greater emphasis on recent observations. This method aligns with scenarios where the latest samples are more pertinent. Tibshirani et al.\cite{tibshirani2019conformal} address a covariate shift by assuming consistent conditional distributions $P_{Y|X}$ despite divergent input distributions between training and testing phases. These approaches, however, may not universally apply to all datasets.
\textbf{Learned weights} 
The seminal work \cite{barber2023conformal} uses pre-defined weights, leading to poor generalization. HopCPT \cite{auer2024conformal} addresses this by using a Modern Hopfield Network (MHN) for similarity-based reweighting, which enhances accuracy by prioritizing past instances that closely resemble current conditions \cite{auer2024conformal,ramsauer2020hopfield}. %By viewing a time series as a collection of regimes with distinct uncertainty characteristics, HopCPT differentially weights calibration points, constructing predictive intervals from exchangeable instances within the same regime.
HopCPT assigns weights in time series forecasting, where each weight \( a_{t,i} \) signifies the relevance of a past time step \( i \) to the current time step \( t \). The weights and corresponding normalized weights are computed as follows:
\begin{equation}
a_{t,i} = \beta m(z_t)^\top W_q W_k m(z_i),
\quad 
\tilde{a}_{t,i} = \frac{e^{a_{t,i}}}{\sum_{j \in M} e^{a_{t,j}}}, \quad \forall i \in M,
\end{equation}
where \( z_i \) is the encoded input for time step \( i \), \( m \) is an encoding function, and \( W_q \), \( W_k \) are matrices of learned weights. The hyperparameter \( \beta \) adjusts the focus of the probabilistic distribution generated by the softmax function.
Using these weights, HopCPT generates the prediction interval by discounting the quantiles of relative errors:
\begin{equation}
\hat{C}_\alpha (x_t, M) = \left[ \hat{f}(X_t) + Q_{\frac{\alpha}{2}} \left( \sum_{i \in M} \tilde{a}_{t,i} \delta_{\epsilon_i} \right), \hat{f}(X_t) + Q_{1 - \frac{\alpha}{2}} \left( \sum_{i \in M} \tilde{a}_{t,i} \delta_{\epsilon_i} \right) \right],
\end{equation}
where \( Q_\tau \) denotes the \( \tau \)-quantile and \( \delta_{\epsilon_i} \) is a point mass distribution of the error $\epsilon_i$ at time step \( i \).
To train the MHN, an auxiliary task using absolute errors as proxy values is employed, aligning errors from comparable regime time steps to predict the current absolute error. 
% \begin{equation}
% \mathcal{L} = T^{-1} \left\| \left( |\epsilon_{1:T}| - A_{1:T,1:T} |\epsilon_{1:T}| \right) \right\|_2^2.
% \end{equation}

%In essence, HopCPT stands at the vanguard of CP techniques, especially for real-world time series datasets, offering unparalleled precision in forecasting\cite{auer2024conformal}.
%Looking ahead, the quest for enhanced CP lies in formulating weight selection mechanisms that are inherently attuned to the temporal dependencies prevalent in time series data. This endeavor remains a critical trajectory for research, with the potential to profoundly impact predictive accuracy in various applications.
\textbf{Discussion} HopCPT exemplifies cutting-edge CP techniques for time series data, offering remarkable forecasting precision \cite{auer2024conformal}. Future research can focus on refining weight selection mechanisms to better capture temporal dependencies in data (e.g., \cite{chen2024conformalized}). Advancing this aspect of CP can lead to significant theoretical innovations and enhance practical forecasting performance.

\subsubsection{Updating non-conformity scores} %\lu{This and Sec 5.1.3 are a bit long. Reduce some.} 
In contrast to reweighting methods that adjust weights based on a predefined strategy, this section introduces updating non-conformity scores to address non-stationary time-series data. By leveraging the most recent $T$ samples and continuously updating prediction intervals as new data becomes available, this approach adapts the intervals to reflect the latest data trends. For example, when data is continuously streamed, updating non-conformity scores uses only the latest data points to calculate prediction intervals, quickly adapting to new trends or shifts in the data. 
% In contrast, reweighting methods typically involve assigning fixed weights to all data points in the calibration set based on predefined criteria, such as similarity to test data or temporal proximity. This can lead to outdated predictions if the underlying data distribution changes significantly. 
Therefore, updating non-conformity scores dynamically integrates new information without segregating data into calibration and test subsets. %This method is ideal for environments with changing statistical properties
 % , requiring a flexible and responsive analytical approach.

A pioneering work proposes EnbPI \cite{xu2021conformal}, which uniquely updates the non-conformity score with sequential error patterns to adapt the intervals dynamically. This approach starts with bootstrap training data to create effective “leave-one-out” (LOO) ensemble prediction models with input \(\hat{X}\):
\(
f_{\hat{X}}(\cdot) = \phi(\{f_{b}(\cdot) : t \not\in S_{b}\}),
\)
where \(\phi\) denotes an arbitrary aggregation function (e.g., mean, median) over a set of scalars, and \(S_{b} \subset [T]\) is the bootstrap index set used to train the \(b\)-th bootstrap estimator \(f_{b}\). The point predictor on test data is defined as \(f_{\hat{X}}(X_{t}) \), which aggregates all bootstrap predictions, reducing the possibility of overfitting.
EnbPI then updates the past residuals during predictions to ensure the prediction intervals have adaptive width. For a fixed \(h \geq 1\), define $\epsilon^h_{t} := \{\hat{\epsilon}_{1}, \ldots, \hat{\epsilon}_{t-h}\}$,
where the residuals are \(\hat{\epsilon}_{t} = Y_{t} - f_{\hat{X}}(X_{t})\).
The prediction intervals \(C_{t-1}(\hat{X})\) are given by:
\begin{equation}
\left[ f_{\hat{X}}(X_{t}) + Q_{\alpha/2}(\epsilon^T_{t}), f_{\hat{X}}(X_{t}) + Q_{1-\alpha/2}(\epsilon^T_{t}) \right],
\end{equation}
which utilize the past \(h = T\) residuals and resemble traditional CP intervals due to the use of the empirical quantile function \(Q_{1-\alpha/2}\) to compute interval width.
This method offers computational efficiency, robustness against overfitting, and adaptability by avoiding the need to retrain models, instead using LOO ensemble models \cite{xu2021conformal}. %This approach efficiently utilizes computational resources and mitigates overfitting, making EnbPI a dynamic tool for real-world applications with changing data distributions, ensuring accurate and current prediction intervals.

The approximately valid coverage guarantee of the EnbPI method is contingent on the assumption that the error terms $\{\epsilon_t\}_{t\geq1}$ are both stationary and strongly mixing.  However, this assumption may not hold in time-series models. SPCI \cite{xu2023conformal} advances beyond EnbPI by relaxing the strict assumptions on error distributions. SPCI incorporates quantile regression to effectively address serial dependencies among residuals in sequential analysis. 
%In particular, SPCI directly leverages the dependency of $\hat{\epsilon}_t$ on the past residuals when constructing the prediction intervals. 
It replaces the empirical quantile with an estimate by a conditional quantile estimator and adjusts the prediction interval dynamically by optimizing its width. This is done by selecting a threshold parameter that minimizes the gap between two conditional quantile estimates while ensuring the desired coverage level.  %Specifically, let $\hat{Q}_t(p)$ be an estimator of the true quantile $Q_t(p)$ and let $\hat{f}$ be a pre-trained point predictor, SPCI intervals $\hat{C}_{t-1}(X_t)$ are defined as
%\begin{equation}
%[\hat{f}(X_t) + \hat{Q}_t(\hat{\beta}), \hat{f}(X_t) + \hat{Q}_t(1 - \alpha + \hat{\beta})],
%\end{equation}
%where $\beta$ minimizes interval width:
%\begin{equation}
%\hat{\beta} = \argmin_{\beta \in [0,\alpha]}(\hat{Q}_t(1 - \alpha + \beta) - \hat{Q}_t(\beta)).
%\end{equation}

%In particular, SPCI is more general than both EnbPI and split conformal. If we train LOO point predictors, choose the quantile estimator $\hat{Q}_t(\cdot)$ as the empirical quantile, and use $\hat{\beta} = \alpha/2$, SPCI reduces to EnbPI If we follow split CP to train the point predictor $\hat{f}$, train quantile predictor $\hat{Q}_t$ on residuals from calibration set, and do no update residuals during prediction, SPCI intervals reduce to the split conformal intervals.

Follow-up studies on SPCI have introduced more sophisticated methods to handle the dependent structure of the time-series data. For example, \citet{lee2024kernel} applies a reweighted Nadaraya-Watson estimator within the realm of nonparametric kernel regression, thus enhancing the selection of data-dependent optimal weights. Additionally, \citet{lee2024transformer} adopts a Transformer decoder as a conditional quantile estimator in the SPCI framework, recognizing its effectiveness in sequential modeling tasks. Both methods facilitate the computation of weights in a data-driven manner, eliminating the need for prior knowledge about the data.

\textbf{Discussion} The "updating" approach for time series is akin to reweighting, but it uniquely utilizes data-adaptive weights. By incorporating the most recent 
$T$ samples into the calibration set, this method ensures that the weights adjust dynamically to reflect the latest data trends and patterns. This similarity to reweighting, yet with a focus on adaptive recalibration, maintains the model's relevance and accuracy amidst evolving data. Future enhancements could involve developing more sophisticated algorithms to determine optimal 
$T$ values are based on real-time changes in data volatility or distribution patterns. Integrating predictive analytics to anticipate the need for updates could further refine its effectiveness.

\subsubsection{Updating the miscoverate rate} Another direction for CP in time series focuses on adaptively adjusting the miscoverate rate $\alpha$ during test time to account for mis-coverage. This method can dynamically adjust the size of prediction sets in an online setting where the data-generating distribution varies over time in an unknown fashion. Such an approach is particularly advantageous for time series data. The continuous recalibration during the testing phase ensures that the prediction intervals maintain an appropriate level of coverage over time.

Adaptive conformal inference (ACI) \cite{gibbs2021adaptive} is designed for constructing prediction intervals for time series data in an online learning framework. %Given a sequence of covariate-response pairs \( (X_t, Y_t) \), the goal is to ensure that the prediction interval \( \hat{C}_t \) covers the true value \( Y_t \) with a probability of at least \( 100(1 - \alpha) \% \). 
The approach is
simple in that it requires only the tracking of a single parameter that models the shift. %and general since it can be combined with any modern ML algorithm that produces point predictions or estimated quantiles for the response.
The update rule for the significance level \( \alpha \) is: $\alpha_{t+1} = \alpha_t + \gamma(\alpha - \text{err}_t), $ where \( \gamma \) is a step size parameter, and \( \text{err}_t \) indicates if \( Y_t \) is excluded from \( \hat{C}_t(\alpha_t) \). The approach ensures that the prediction intervals adjust over time to account for shifts in the data distribution, maintaining the desired coverage probability.
% \textbf{Adaptive selection for $\gamma$} 

The choice of $\gamma$ gives a tradeoff between adaptability and stability. While raising the value of $\gamma$ will make the method more adaptive to distribution shifts, it will induce greater volatility in $\alpha_t$. 
\citet{zaffran2022adaptive} theoretically analyzed the influence of $\gamma$ on ACI efficiency and proposes AgACI which runs multiple ACI instances with different learning rates and combines their outputs using an online aggregation of experts approach. Alternatively, \citet{gibbs2022conformal} dynamically adjusts \(\gamma\) based on expert performance. Each expert operates with a specific \(\gamma\), and ACI is performed in parallel for all experts. The final \(\alpha_t\) is selected based on the historical performance of each expert, effectively learning the optimal \(\gamma\) over time to adapt to shifts in the data distribution.

Another alternative to ACI is the multi-valid CP (MVP) method of \cite{bastani2022practical} which accounts for worst-case adversarial guarantees to ensure the methods remain reliable even when data distributions are manipulated or shift unexpectedly, especially in adversarial environments.
MVP achieves this by selecting prediction thresholds that control the long-term miscoverage rate. %Instead of targeting a single optimal parameter, MVP dynamically adjusts to adversarial conditions, maintaining robust prediction intervals across varying thresholds.
Similarly,  \citet{angelopoulos2024online} address worst-case scenarios by examining the effect of a decaying step size on the prediction threshold update rule. Their approach ensures strong long-term coverage in adversarial settings and achieves convergence to an optimal threshold.% in independent and identically distributed (I.I.D.) scenarios.

Another follow-up work of ACI focuses on refining the regret condition within the online learning framework. SAOCP \cite{bastani2022practical} targets minimizing \textit{Strongly Adaptive Regret} (SAReg), which evaluates performance over all intervals of specific lengths %for finer adaptation to data shifts, 
rather than over the entire length. %SAOCP uses a collection of online learning experts, each tuned for different intervals and dynamically weighted by performance, ensuring predictions are driven by the most effective expert. Predictions are adjusted via a modified gradient descent, enhancing adaptability to evolving data. SAOCP outperforms standard regret-based methods by employing SAReg to dynamically adapt performance across varied intervals, ensuring high precision in changing environments. It integrates Scale-Free Online Gradient Descent (SF-OGD) for adaptable learning rates and uses expert algorithms optimized for specific data segments, ensuring robust and accurate predictions. 
%This approach offers a significant performance advantage over ACI and DtACI.
In \cite{angelopoulos2024conformal} , control theory is used to dynamically manage uncertainties and adapt to changing data. It integrates ACI with traditional PID control mechanisms to enhance predictions. Additionally, \citet{yang2024bellman} proposes BCI to addresse ACI’s limitations by incorporating Model Predictive Control (MPC) techniques to optimize prediction intervals. %BCI proactively adjusts prediction intervals to optimize their efficiency and coverage using a stochastic control framework and dynamic programming. 
Together, these advancements extend ACI’s capabilities by enhancing adaptability and robustness in dynamic and adversarial environments.

\textbf{Discussion} Approaches that dynamically adjust the miscoverate rate based on incoming data, akin to how online learning updates model parameters in response to new information. Future research could explore more nuanced methods for adjusting miscoverate rate, potentially incorporating reinforcement learning to optimize decision-making processes based on performance feedback. Importantly, these methods can be synergistically combined; for instance, updating may be viewed as a specialized form of reweighting that places greater emphasis on recent samples. Moreover, the integration of reweighting with adaptive strategies has been shown to enhance performance significantly \cite{auer2024conformal,hajibabaee2024adaptive}. While the strategies outlined are broadly applicable to spatio-temporal data, research exists for specific types of time series, such as AR process \cite{zaffran2022adaptive}, Hidden Markov Model \cite{nettasinghe2023extending}, and $\beta$-mixing process \cite{oliveira2022split}, which further refine these general techniques.

\subsubsection{Other methods}
\citet{chernozhukov2018exact} develops a randomization method and accounts for potential serial dependence by including block structures in the permutation scheme, which can be either overlapping or non-overlapping. 
The non-overlapping blocking scheme divides the data into separate blocks to reduce dependencies within each block. In contrast, the overlapping blocking scheme shifts the starting point of each block slightly to create overlapping sections between them. This overlap allows the scheme to capture more variations and dependencies between observations by including samples that are close together in multiple blocks. %These methods help CP maintain valid prediction intervals by effectively accounting for data dependencies, resulting in more reliable predictions when samples are dependent.
 Follow-up work such as \cite{ajroldi2023conformal} employs similar blocking techniques to account for dependencies in a probabilistic forecasting scheme for two-dimensional functional time series.
 \citet{cleaveland2024conformal} propose an optimization-based method that reduces the conservatism of CP for time series predictions. Instead of treating prediction errors independently at each time step, they consider a parameterized prediction error over multiple time steps and optimize the parameters over an additional dataset.  \citet{prinzhorn2024conformal} further analyzes different temporal components using time series decomposition. This method decomposes the original time series into remainder, trend, and seasonal components %, which are classified as globally exchangeable, non-exchangeable, and locally exchangeable, respectively. They 
and leverage previous methods such as EnbPI \cite{xu2021conformal} and ACI \cite{gibbs2021adaptive} to handle these different components. This strategy better captures the inherent characteristics of time series and improves performance.

A very recent work by \cite{oliveira2024split} extends split CP to handle non-exchangeable data by introducing a novel framework based on concentration inequalities and data decoupling properties. The method assumes that the calibration and test data distributions are reasonably similar and stable, ensuring reliable prediction intervals with guarantees for both marginal coverage and conditional coverage. %By incorporating small penalties to account for dependencies, the approach effectively handles stationary processes such as ARMA models or Markov chains, where data points exhibit temporal or spatial correlations.

\subsection{Multi-Dimensional}
Multi-dimensional spatio-temporal data spans multiple dimensions of both time and space, introducing additional complexity and opportunities for predictive modeling. The dependencies between spatial and temporal dimensions further undermine the exchangeability assumption inherent in traditional methods. Previous works on multi-dimensional data primarily focus on exchangeable data \cite{johnstone2021conformal,messoudi2022ellipsoidal,feldman2023calibrated,ghosh2023probabilistically}, overlooking the temporal and spatial dependencies. This section mainly focuses on advanced techniques that adapt CP for multi-dimensional spatio-temporal data in regression problems. %By leveraging the intricate patterns and interconnections in multi-dimensional datasets, we aim to enhance the accuracy and applicability of CP in these complex contexts.

%\subsubsection{Regression} In this section, we investigate research on multi-dimensional time series forecasting frameworks. 
A simple approach is
to treat each time step as a separate prediction task. CF-RNN \cite{stankeviciute2021conformal} constructs prediction intervals using Bonferroni correction, selecting the 
$(1-\alpha/k$)-th quantile for a confidence level of 
$1-\alpha$. This method results in conservative intervals and ignores spatial dependencies. Similarly, \citet{mao2024valid} achieves theoretical validity by assuming that spatial locations are exchangeable. Without making this assumption, they focus on specific locations and propose a localized approach that utilizes nearby locations to address spatial dependence. \citet{zhou2024conformalized} also assume that each time series trajectory is exchangeable, combining with ACI to produce more informative prediction intervals for motion planning.
%We'll first focus on longitudinal coverage, which addresses the accuracy of predictions over time for individual series. Then, we will explore cross-sectional coverage, which ensures predictions are accurate across different series at specific time points. For example, in forecasting patient data over days in a hospital, cross-sectional validity confirms that predictions are accurate across all patients at specific time points.
%We denote our time series dataset by $\mathcal{D} = \{ (x_{1:t}^{(i)}, y_{t+1:t+k}^{(i)}) \}_{i=1}^{n}$, where $x_{1:t} \in \mathbb{X}^t$ represents the $t$ timesteps of input, and $y_{t+1:t+k} \in \mathbb{Y}^k$ indicates the $k$ timesteps of output. Traditionally, in time series forecasting, the input and output spaces ($\mathbb{X}$ and $\mathbb{Y}$) are equivalent, though they do not necessarily have to coincide.
%Given the dataset $\mathcal{D}$, a test sample $x_{1:t}^{(n+1)}$, and a confidence level $1-\alpha$, our goal is to establish a set of $k$ prediction intervals $\Gamma_{t+1}^{1-\alpha}, \dots, \Gamma_{t+k}^{1-\alpha}$ for each timestep, such that:
%\[
%\Pr\left( \forall h \in \{1, \dots, k\}, y_{t+h}^{(n+1)} \in \Gamma_{t+h}^{1-\alpha} \right) \geq 1 - \alpha
%\]
%This approach aligns with the objectives of CP, enabling the generation of reliable prediction intervals under assumptions of data distribution shifts, thereby providing robust forecasts for multivariate time series.
 \citet{sun2023copula} considers spatial dependence by using copulas to model uncertainty over future time steps, reducing confidence interval widths while maintaining validity. Copulas, which model dependencies between variables, apply the empirical cumulative distribution function (CDF) to estimate nonconformity scores, ensuring the predicted trajectory falls within the confidence intervals at a level of 
\(1 - \alpha\).

%Adjustments in significance levels \(\alpha\) at each step are managed using an empirical copula, \(C_{\text{empirical}}\), which is calculated based on the cumulative probabilities of each sample. To ensure the validity of CP, the cumulative probabilities \(\textbf{u}^*\) must satisfy \(C_{\text{empirical}}(\textbf{u}^*) \geq 1 - \alpha\). 
%Optimal values are found using a suitable search algorithm, and the prediction region for each timestep includes all possible outcomes where the nonconformity score does not exceed a certain threshold.
%\[
%F(s) = \frac{1}{|D_{cal-1}|} \sum_{z^i \in D_{cal-1}} 1_{s(z^i,f) \leq s}.
%\]

%For multi-step predictions, it calculates an empirical CDF for each step, \(\hat{F}_t(s_t) = P(S(z_t, f) \leq s_t)\), ensuring the predicted trajectory falls within the confidence intervals at a level of \(1 - \alpha\). Adjustments in significance levels \(\alpha\) at each step are managed using an empirical copula \(C_{\text{empirical}}(\textbf{u})\):

%\[
%C_{\text{empirical}}(\textbf{u}) = \frac{1}{|D_{cal-2}|} \sum_{i \in D_{cal-2}} 1_{\textbf{u}^i \leq \textbf{u}},
%\]

%where \(\textbf{u}\) denotes the cumulative probabilities of each sample. To ensure CP's validity condition:

%\[
%\textbf{u}^* = (\hat{F}_1(s_1^*), \ldots, \hat{F}_k(s_k^*)),
%\]
%\[
%C_{\text{empirical}}(\textbf{u}^*) \geq 1 - \alpha,
%\]
 %This method effectively addresses distribution shifts and maintains accurate predictions over multiple steps.

The theory behind \cite{sun2023copula} assumes that each data sample in a time series is i.i.d., thus ignoring temporal dependencies. \citet{xu2024conformal} relax this assumption by constructing episodic prediction sets that are adaptively and efficiently calibrated during test time to account for temporal dependencies.
%The theory of CopulaCPTS assumed that each data sample of an entire time series is drawn i.i.d. from an unkonwn distribution, those ignoring temporal dependency. \cite{xu2024conformal} breaks this assumption and build eppiposidal prediction sets whose size is adpatively and efficiently calibrated during test time. %In particular, sizes of the ellipsoids are sequentially re-estimated during test time to ensure adaptiveness and accuracy.
This method starts with a training dataset \( Z_t = (X_t, Y_t) \) for \( t = 1, \ldots, T \), and a prediction algorithm \( \hat{f} \) that forecasts \( Y \). Key to this method is the computation of prediction residuals \( \hat{\sigma}_t = Y_t - \hat{f}(X_t) \), which are used to estimate the covariance \( \hat{\Sigma} \), because it captures the dependencies and relationships between different dimensions of the data. The covariance is:
% estimated as:
\vspace{0pt}
\begin{equation}
\hat{\Sigma} = \frac{1}{T - 1} \sum_{t=1}^{T}(\hat{\sigma}_t - \bar{\sigma})(\hat{\sigma}_t - \bar{\sigma})^T,
\end{equation}
\vspace{0pt}
where \( \bar{\sigma} \) is the mean of \( \hat{\sigma}_t \). 
They ensure the invertibility of \( \hat{\Sigma} \) via a low-rank approximation, crucial for defining the ellipsoidal (prediction) region \( B(r, \bar{\sigma}, \hat{\Sigma}_{\rho}) \) with a radius \( r \). This radius is calibrated through CP by calculating a non-conformity score for a new sample \( (X, Y) \):
\begin{equation}
\hat{e}(Y) = (\hat{\sigma} - \bar{\sigma})^T\hat{\Sigma}_{\rho}^{-1}(\hat{\sigma} - \bar{\sigma}),
\end{equation}
\noindent where \( \hat{\sigma} = Y - \hat{f}(X) \). The method then determines \( r \) to ensure predictions fall within the ellipsoid at a specified confidence level. The prediction sets are then constructed via SPCI \cite{xu2024conformal,xu2023sequential}.
%Follow SPCI framework, the corresponding the prediction set \( \hat{C}_{t-1}(X_t, \alpha) \subseteq \mathbb{R}^p \) for a given confidence level \( \alpha \) takes the form
%\begin{align*}
%\hat{C}_{t-1}(X_t, \alpha) &= \{ Y: \hat{Q}_t(\beta) \leq \hat{e}(Y) \leq \hat{Q}_t(1 - \alpha + \beta) \}\\
%&= \{ \hat{f}(X_t) + B(\hat{Q}_t(1 - \alpha + \beta), \hat{e}, \hat{\Sigma}_{\rho}) \ | \ B(\hat{Q}_t(\beta), \hat{e}, \hat{\Sigma}_{\rho}) \},
%\[
%\hat{\beta} = \argmin_{\beta \in [0,\alpha]} V(\hat{\Sigma}_{\rho}, \hat{Q}_t(1 - \alpha + \beta)) - V(\hat{\Sigma}_{\rho}, \hat{Q}_t(\beta)).
%\]

\subsubsection{Cross-sectional Coverage}
In the multi-dimensional time series domain, an interesting branch considers both cross-sectional and longitudinal validity simultaneously. %For example, in predicting patient data over time, such as days in a hospital, cross-sectional validity ensures that predictions are accurate across all patients at specific time points, while longitudinal validity guarantees consistent accuracy over time for individual patients. 
Two works \cite{lin2022conformala,lin2022conformalb} enhance prediction intervals in time series forecasting by ensuring both cross-sectional and longitudinal coverage. \citet{lin2022conformala} introduce Temporal Quantile Adjustment (TQA), which dynamically adjusts quantiles using temporal information. TQA employs a two-step strategy: predicting future quantiles based on nonconformity scores and refining them through a budgeting strategy (TQA-B) or error-based adjustment (TQA-E). By contrast, \citet{lin2022conformalb} focus on normalization to account for distribution changes. It calculates these scores based on historical data discrepancies and normalizes them with the median absolute deviation to adjust for shifts in data distribution over time.

\subsubsection{Classificaion and anomaly detection}
Classification and anomaly detection within spatio-temporal data contexts also hold significant importance. %Accurate forecasts of such events can profoundly impact public safety by enabling better preparation and minimizing potential damages. 
Building upon their foundational research, \citet{xu2022conformal} developed the Ensemble Regularized Adaptive Prediction Set (ERAPS), which adapts the EnbPI framework 
 \cite{xu2021conformal} for classification problems by applying the Regularized Adaptive Prediction Set (RAPS) method \cite{angelopoulos2020uncertainty} from image classification to construct non-conformity scores. Additionally, \cite{xu2021conformala} further developed EnbPI for anomaly detection in multi-dimensional data with ensembling to obviate the need for data splitting% by adapting the LOO ensemble predictor
 and enhance prediction accuracy and robustness.
%Similar to EnbPI introduced by \cite{xu2021conformal}, ERAPS utilizes ensembling obviate the need for data splitting by adapting the leave-one-out (LOO) ensemble predictor to enhance prediction accuracy and robustness. This methodology has been effectively applied to predict fire sizes in various regions \cite{xu2022wildfire}, demonstrating its practical utility and reliability in real-world scenarios.

%\textbf{Anomaly Detection}
%Anomaly detection is crucial for spatio-temporal data, such as identifying anomalous traffic flows. Ensemble Conformal Anomaly Detection (ECAD) \cite{xu2021conformala} is the first conformal anomaly detector for spatio-temporal observations, providing approximately valid theoretical guarantees. ECAD uses residuals from ensemble regression models as anomaly scores and performs detection through local comparisons. It employs a two-phase approach: the training phase computes anomaly scores based on absolute residuals and generates multiple bootstrap non-conformity mappings, which are aggregated using statistical methods to refine anomaly scores. In the detection phase, these scores are compared to a threshold set by the $1-\alpha$ quantile of the bootstrap predictions to determine anomalies. The method dynamically updates the sequence of anomaly scores, accommodating new data and discarding old scores as needed.

\textbf{Discussion} This section discusses methods for handling dependencies in time and space, highlighting the challenge of solving non-exchangeability issues. Additionally, simultaneously considering longitudinal and cross-sectional coverage is an area that requires further exploration. Finally, the use of graphical models for multi-dimensional time series remains underexplored, presenting a promising direction for future research.

\subsection{Streaming Data}
Streaming data is characterized by continuous flow, high volume, and rapid updates. Adaptive CP(ACP) methods \cite{gibbs2021adaptive, gibbs2022conformal} for streaming data dynamically adjust $\alpha$ to maintain desired miscoverage rate as data characteristics evolve.
In this section, we first introduce the application of ACP in motion planning. Next, we discuss the use of CP for detecting concept drift and enhancing interpretability in streaming data and finally extend CP to online learning beyond numerical data streams.

\subsubsection{Streaming numerical data}

%\subsubsection{Motion Planning}

Motion planning involves navigating agents from an origin to a destination without collisions, despite uncertainties from environmental factors, sensor errors, and model inaccuracies. The environment includes \( N \) dynamic agents with states \( Y_t \) following an unknown distribution \(\mathcal{D}\). ACP manages streaming data to forecast future states over a prediction horizon \( H \), updating the adaptive threshold \(\delta_{\tau}\) based on prediction accuracy:
\begin{equation}
\delta_{\tau+1} := \delta_{\tau} + \gamma(\delta - e_{\tau}),
\end{equation}
where \( e_{\tau} \) indicates prediction accuracy, and \( \gamma \) is a tuning parameter.
Using ACP, an Uncertainty-Informed Model Predictive Controller (MPC) optimizes a control sequence to ensure safe navigation by integrating real-time data and adapting to dynamic environments \cite{dixit2023adaptive, lindemann2023safe}.

CP can also help monitor online data streams, particularly in tasks like concept drift detection \cite{tanha2022cpssds,ma2023metastnet,eliades2021using} and regression trees \cite{johansson2014regression,johansson2018interpretable,johansson2019interpretable}. For instance, Inductive CP (ICP) assigns non-conformity scores to new samples based on a model trained on initial data. If the p-value for a new sample falls below a certain threshold, indicating a potential concept drift, a new classifier is trained; otherwise, the model is updated with the current labeled data. Additionally, CP enhances both interpretability and efficiency in regression trees by applying a Mondrian conformal predictor at each leaf, focusing on localized updates to maintain accuracy and adapt to changes in the data stream.

%\subsubsection{Concept Drift Detection}

%CP monitors online data streams for concept drift, critical in semi-supervised classification tasks where limited labeled data accompanies a larger volume of unlabeled data. In Inductive CP (ICP), a model \( H \) trained on initial samples assigns non-conformity scores to new samples. The p-value for a new sample \( x \) is:
%\[
%p_y^{x_i} = \frac{|\{i = 1, \ldots, (n+1) : \alpha_x^i \geq \alpha_{x_{n+1}}^y\}|}{n + 1}.
%\]
%A p-value below a threshold \( \epsilon \) indicates significant deviation, suggesting concept drift. If drift is detected, a new classifier is trained; otherwise, the existing model is updated with the current chunk's labeled data. This dynamic approach ensures model accuracy and reliability in changing environments.

%\subsubsection{Regression Trees for Data Streams}

%Managing regression trees in online data streams is challenging, particularly for maintaining interpretability and computational efficiency. CP enhances interpretability by ensuring statistical validity from root to leaves, facilitating reliable decision-making. The Global Online Inductive Conformal Predictor (G-OICP) develops a regression tree from initial training data and updates nonconformity scores with new data. The Local Online Inductive Conformal Predictor (L-OICP) applies a rule-conditioned Mondrian conformal predictor to each leaf, localizing updates and preserving accuracy.

\subsubsection{Beyond Numerical Data}

Other studied problems with streaming data include image classification and document retrieval in online learning. %Online CP with semi-bandit feedback constructs prediction sets based on a scoring function, ensuring true label coverage with confidence \( \alpha \). The algorithm observes new data, updates the empirical CDF, and adjusts thresholds using the Dvoretzky-Kiefer-Wolfowitz (DKW) inequality:
%\begin{equation}
%    \overline{G}_t(\tau) = G_t(\tau) + \epsilon_t,
%\end{equation}
%where \( \epsilon_t = \sqrt{\frac{\log(2/\delta)}{2t}} \). This process maintains valid thresholds, ensuring accuracy in real-time data streams without full retraining, thus optimizing computational efficiency. The algorithm achieves sublinear regret compared to the optimal conformal predictor, maintaining accuracy over time.
\citet{ge2024stochastic} explore the use of stochastic online CP with semi-bandit feedback in real-time data streams, where the system only receives information about the true label if it is included in the prediction set. %For example, in a document retrieval task, a user confirms the correctness of a retrieved document only if it matches the target within the predicted set. 
Stochastic online CP updates its prediction thresholds using the Dvoretzky-Kiefer-Wolfowitz (DKW) inequality to ensure the empirical distribution of non-conformity scores up to time $t$ is not overly optimistic about the true distribution of the data by adding an adjustment factor. The adjusted threshold ensures that the prediction intervals remain conservative enough to maintain the desired coverage.
\\
\noindent\textbf{Discussion} The seminal work ACI can be a general strategy to handle high-volume, non-stationary streaming data. When applying CP for streaming data, it is crucial to consider computational efficiency. Additionally, there is a significant need for more research on scalability, particularly in distributed systems where data streams are processed in parallel across multiple nodes. Developing efficient algorithms that can maintain statistical guarantees while operating in such environments would be highly beneficial. Finally, exploring the application of CP in multi-modal data streams, where data comes from various sources and types, presents another exciting research frontier.

\section{Open challenges and future directions}
% We outline in this section open challenges and potential future directions for CP.
\textbf{Non-exchangeability: distribution heterogeneity and temporal dependence.} 
The exchangeability assumption often fails in real-world scenarios. While existing methods like non-exchangeable CP \cite{barber2023conformal} assess the coverage gap of weighted CP based on the distance between distributions, developing an adaptive approach to minimize this gap remains an open challenge. Moreover, works such as ACI and related studies \cite{gibbs2021adaptive,angelopoulos2024online} provide adversarial coverage guarantees for sequences with arbitrary distributions, but they do not ensure validity at each step. Given this, it would be valuable to explore how light distributional assumptions on the sequence could enhance adaptivity.

% In many practical scenarios, such as time series and natural language data, the assumption of exchangeability is unrealistic. Specifically, quantifying the coverage gap between exchangeable and non-exchangeable data is a significant challenge. Furthermore, choosing appropriate data-adaptive weights to balance the distribution between calibration and test data is crucial.

\textbf{Robust CP with more general data imperfections.} 
Ensuring the reliability of CP methods under data imperfections and adversarial conditions is crucial. As highlighted in the survey, critical scenarios include observations with missing values, noisy entries, or adversarial attacks. The validity of some existing methods hinges on specific distributional assumptions, such as the missing at-random assumption. However, real-world missingness can follow more complex patterns, like block-wise or non-random missingness. Extending CP to handle such general data imperfections, while maintaining prediction set validity with tractable coverage gaps, remains an open problem.

% One key area is handling missing values, where techniques are needed to effectively manage incomplete data without compromising the validity of predictions. Another challenge is dealing with noisy labels, which occur when the training data contains incorrect labels. Moreover, defending against adversarial attacks where data may be deliberately manipulated is critical.

\textbf{Handling emerging data types.} 
As data evolves, new data types bring challenges and opportunities for CP. For example, multi-modal data, which integrates information from various sources such as text, images, and sensor data, presents the challenge of combining heterogeneous data sources with differing noise characteristics and distributions. A key open question is how to "optimally" aggregate data from multiple modalities to enhance CP performance. Similarly, the rise of streaming data underscores the need for adaptive and online CP techniques that can provide valid predictions in real time. This requires maintaining computational efficiency and prediction validity as new data arrives and as the underlying data distribution potentially shifts over time.

% \textbf{CP with other frameworks.}
% The integration of CP with other predictive frameworks holds great promise for enhancing predictive performance and reliability. One potential avenue is the combination of CP with reinforcement learning and control theory \cite{yang2024bellman,angelopoulos2024conformal}. This integration can help develop strategies that are not only optimal but also reliable and interpretable. Additionally, the synergy between CP and Bayesian methods can significantly enhance the calibration and UQ of predictive models \cite{cha2023temperature}.

\textbf{CP for responsible AI.}
CP offers a promising approach to responsible AI \cite{lu2022improving}. CP can generate clear, quantifiable confidence measures, aiding users in understanding and trusting model predictions, which promotes interpretability \cite{johansson2018interpretable,johansson2019interpretable,huang2024confine}. For fairness, CP can help equalize coverage across diverse population groups, correcting biases without the need for expensive retraining \cite{romano2020classification,lu2022fair,wang2023equal}. Privacy is another critical area, especially in federated learning \cite{lu2023federated,lu2021distribution}; CP can be adapted to produce confidence intervals without accessing sensitive or personally identifiable information, thus preserving privacy while ensuring reliable predictions. By addressing these aspects, CP can significantly contribute to the development of AI systems that are not only accurate but also fair, transparent, and resilient across various applications.

\textbf{CP in aiding decision-making.}
Research indicates that when humans are provided with CP sets as \emph{side information}, their task accuracy improves compared to using fixed-size prediction sets, thereby enhancing human-in-the-loop decision-making and human-AI collaboration \cite{cresswell2024conformal}. CP also aids in early stopping for iterative algorithms by providing robust uncertainty measures, which helps prevent overfitting and reduces computational costs \cite{liang2023conformal}. In online auctions, CP can predict final prices and dynamically adjust bidding strategies, leading to improved efficiency and profitability \cite{han2024conformal}. These applications highlight CP's potential to transform various fields through reliable and adaptive predictions. Thus, it is essential to explore additional scenarios where CP can be applied to further unlock its potential and broaden its impact.

\textbf{CP for cross-disciplinary research.}
Cross-disciplinary research offers significant opportunities to advance CP. For example, insights from statistics can contribute to developing new theoretical foundations and methods for CP. Collaborations with domain experts in areas like healthcare, finance, and environmental science can lead to the creation of application-specific CP methods that address unique challenges in these fields. Additionally, leveraging advances in computer science, particularly in optimization and high-performance computing, can improve the scalability and efficiency of CP techniques. Interdisciplinary research that bridges theory and practice is also crucial for translating CP methods into real-world applications, thereby maximizing their impact.

\textbf{Open source tools for CP.} Despite progress in developing open-source tools for CP, challenges remain in expanding their usability. Current tools like PUNCC \cite{mendil2023puncc} and TorchCP \cite{wei2024torchcp} address traditional tasks such as regression, classification, and graph-based learning, while Fortuna \cite{detommaso2024fortuna} takes a broader approach, incorporating uncertainty quantification techniques such as Bayesian inference. Additionally, \citet{susmann2023adaptiveconformal} introduced an open-source R package for ACI. 
While many tools effectively handle traditional data types, there is a notable lack of support for modern data types, such as unstructured data and streaming data, and for addressing the rapid growth of language modeling applications. %Current tools like MAPIE \cite{mapie} lack support for modern data types and integration with deep learning frameworks, limiting their use in large-scale, complex scenarios. Additionally, there is limited support for streaming data, which is increasingly needed for dynamic environments.
Improving compatibility, scalability, and ease of use across machine learning platforms is crucial to foster wider adoption and innovation in CP.

\bibliographystyle{plainnat}
\bibliography{main}
\appendix
\section{Appendix1}
\begin{table}[htb]
\small
\centering
\begin{tabular}{|l|l|}
\hline
\textbf{Notation} & \textbf{Description} \\ \hline
$X_i, \mathcal{X}$ & feature of $i$-th instance and feature domain\\ \hline
$Y_i, \mathcal{Y}$ & label of $i$-th instance and label domain\\ \hline
$V(X_i, Y_i), V_i$ & nonconformity score of $i$-th instance \\ \hline
$Q_{1-\alpha}, \hat{Q}_{1-\alpha}$ & quantile or estimated quantile at $1-\alpha$ \\ \hline
$\alpha$ & miscoverage rate \\ \hline
$C_{1-\alpha}(X_i)$ & predicted set of $i$-th instance \\ \hline
$D, n$ & dataset and its size \\ \hline
$\sigma_i$ & residuals or estimated residual of $i$-th instance \\ \hline
$P, P_X, P_{Y|X}$ & distribution, distribution of $X$, distribution of $Y$ given $X$\\ \hline
% $k, K$ & class and domain of class \\ \hline
$\hat{F}(X_i, Y_i)$ & estimated CDF \\ \hline
$\pi$ & permutation \\ \hline
$\theta_Z$ & parameter of model get from data $Z$ \\ \hline
$z_i, Z$ & pair of label and feature, equal to $(X_i,Y_i)$ or $(X,Y)$ \\ \hline
$\ell$ & loss function \\ \hline
$\hat{\tau}$ & calibration threshold \\ \hline
$t, T$ & time stamp \\ \hline
$f, g$ & neural network or predictor \\ \hline
$\hat{l}, \hat{u}$ & lower and upper bound of prediction interval\\ \hline
$w_i$ & weight for $i$-th instance \\ \hline
$\delta_{R_i}$ & point mass of $i$-th instance \\ \hline
% $\lambda, \eta, \beta ...$ & hyperparameter \\ \hline
\end{tabular}
\caption{Major notations used throughout the survey.} 
\label{table:your_label}
\end{table}

% \section{Examples}
% \lu{we cannot use images from other papers without getting authors' consent. We don't have to include them.}
% \textbf{Example 1}: CP in image segmentation, \cite{mossina2024conformal}

% \begin{figure}[!ht]
%     \centering
%     \includegraphics[width=1\linewidth]{figure/ade20k_miscov_0.01.png}
%     \caption{An example of CP application for image segmentation in \cite{mossina2024conformal}, the heatmap represents the number of potential classes for each pixel}
%     \label{fig:example}
% \end{figure}
% \textbf{Example 2}: In image generation

% \begin{figure}[!ht]
%     \centering
%     \includegraphics[width=1\linewidth]{figure/teaser_CR.pdf}
%     \caption{An example of CP application for image generation in \cite{sankaranarayanan2022semantic}, CP algorithm generates multiple results to cover ground truth.}
%     \label{fig:example2}
% \end{figure}

\end{document}